\documentclass[10pt,twocolumn,letterpaper]{article}

\usepackage[algorithms]{wacv}      

\usepackage{amsmath}
\usepackage{amssymb}
\usepackage{booktabs}
\usepackage{multirow}
\usepackage{graphicx}
\usepackage{xcolor}
\usepackage{enumitem}
\usepackage{algorithm}
\usepackage{algorithmic}
\usepackage{makecell}

\newcommand{\method}{GeoBridge}
\newcommand{\img}{\mathbf{x}}
\newcommand{\coord}{\mathbf{y}}
\newcommand{\cond}{\mathbf{c}}
\newcommand{\tokens}{\mathbf{H}}
\newcommand{\Sph}{\mathbb{S}^{2}}
\newcommand{\dist}{d_{\mathrm{geo}}}

\definecolor{wacvblue}{rgb}{0.21,0.49,0.74}
\usepackage[pagebackref,breaklinks,colorlinks,allcolors=wacvblue]{hyperref}
\usepackage[table]{xcolor}
\definecolor{methodyellow}{RGB}{255,248,220}
\definecolor{baselinegray}{RGB}{238,238,238}

\def\wacvPaperID{2091}
\def\confName{WACV}
\def\confYear{2027}

\title{GeoBridge: Decoupled Semantic Conditioning for Generative Image Geolocalization}

\author{
Zhiyang Dou\textsuperscript{1,$\star$} \quad 
Xumeng Han\textsuperscript{2,$\star$} \quad
Fengde Peng\textsuperscript{3} \quad
Zipeng Wang\textsuperscript{1} \\
Moxuan Zhao\textsuperscript{2} \quad
Zhipei Huang\textsuperscript{2} \quad
Zhenjun Han\textsuperscript{2,$\dagger$}\\
\normalsize
University of Chinese Academy of Sciences \\
\normalsize
\textsuperscript{1}School of Advanced Interdisciplinary Sciences,  \textsuperscript{2}School of Electronic, Electrical and Communication Engineering\\
\normalsize
\textsuperscript{3}Shenzhen Institutes of Advanced Technology\\
}

\begin{document}
\maketitle
\begingroup
\renewcommand{\thefootnote}{\fnsymbol{footnote}}
\footnotetext[0]{%
$\star$ Equal contribution.
$\dagger$ Corresponding author
(\href{mailto:hanzhj@ucas.ac.cn}{hanzhj@ucas.ac.cn}).}
\endgroup

\begin{abstract}
Multimodal large language models (MLLMs) have advanced image geolocalization mainly by improving how they \emph{reason} about geographic cues. How that reasoning is \emph{decoded} into coordinates, however, has lagged behind. Predicting a place name for a geocoding API is discrete and lossy: it ignores image evidence and collapses multi-granular semantics into a coarse lookup. We argue that the bottleneck has shifted from \emph{what} a model reasons to \emph{how} that reasoning is represented for a continuous, geometry-aware decoder. We present \method, a role-decoupled conditioning mechanism that connects a frozen semantic MLLM to a frozen Riemannian flow-matching head that generates coordinates on the sphere. The central obstacle is a \emph{role conflict}: supervising the condition with discrete semantic labels biases its representation toward class-discriminative geometry, at odds with the smooth manifold the generative head requires. \method\ keeps the semantic supervision decoupled from the condition interface: a separate projection forms the continuous condition the frozen head expects, injecting geographic priors without disturbing the spherical decoder. On IM2GPS3K, \method\ reaches $38.67/52.89/70.37$ at the $25/200/750$\,km thresholds, improving over a place-name-to-API pipeline and reasoning-augmented direct prediction at these precision-relevant scales. \method\ is a decode-side algorithmic contribution, orthogonal and complementary to chain-of-thought reasoning. Code will be made publicly available.
\end{abstract}
\section{Introduction}

Worldwide image geolocalization asks a model to predict where a photograph was taken on the Earth~\cite{Im2gps, vo2017revisiting, shen_2017_streetvizor,weyand2016planet}. The evidence in a single image varies sharply: a visible street sign may pin down a city block, while a rural road or coastline supports only a broad regional hypothesis~\cite{OSV5M}. Recent multimodal large language model (MLLM) geolocation systems~\cite{GeoReasoner, gaga, geo-r, globe, gre, geoagent} have advanced this task mainly by improving \emph{reasoning}---reading text, architecture, road layout, and vegetation to infer a plausible place. Far less attention has gone to the step that turns those semantics into an actual coordinate.
 
\begin{figure}[t]
  \centering
  \includegraphics[width=\linewidth]{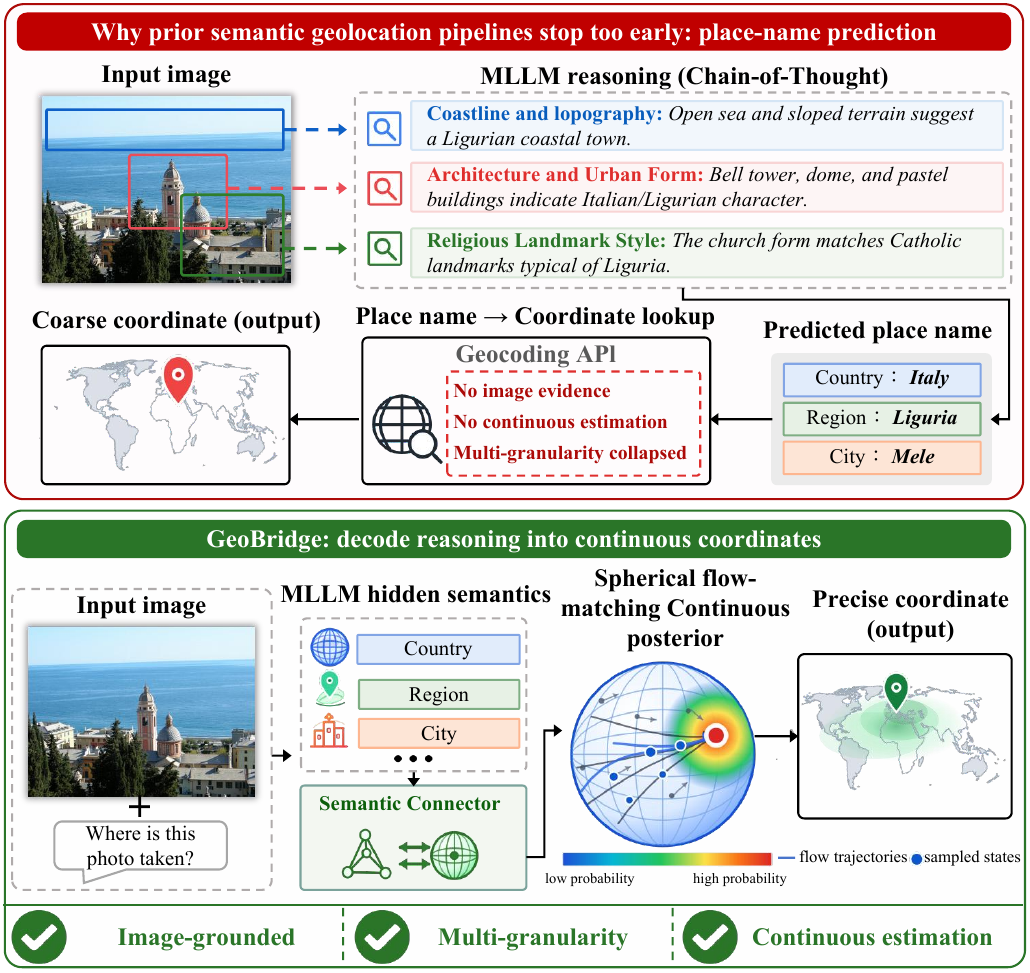}
  \caption{\textbf{From discrete place names to continuous spherical decoding.}
  \emph{Top:} prior semantic geolocation pipelines stop too early. An MLLM reasons over visual cues, but its output is a discrete place name handed to a blind geocoding API.
  \emph{Bottom:} \method\ instead decodes the MLLM's multi-granular hidden semantics through a lightweight semantic connector into a frozen spherical flow-matching head, producing a fine-grained coordinate.}
  \label{fig:pipeline}
\end{figure}

This decode step is where current pipelines stop too early (Fig.~\ref{fig:pipeline}). A common route predicts a place name and calls a geocoding API~\cite{globe,GeoReasoner,gaea}, reducing a fine-grained spatial decision to a discrete database lookup that discards the image evidence and any sub-region detail. Coordinates, however, live on a continuous sphere, where a geometry-aware generative decoder---such as a Riemannian flow-matching (RFM) head~\cite{around_the_world_in_80_timesteps,lipman2022flow}---can place predictions anywhere in continuous space. We argue that the bottleneck has shifted from \emph{what} an MLLM reasons to \emph{how} that reasoning is represented for such a decoder.
 
We therefore study the interface itself: a frozen MLLM that supplies multi-granular geographic semantics, and a frozen RFM head that generates coordinates on the sphere. The central difficulty is a \emph{role conflict}. Discrete semantic labels are valuable priors, but supervising the condition representation directly with them pulls it toward class-discriminative geometry, while the frozen flow head expects a smooth condition compatible with its coordinate manifold.
 
\method\ resolves this conflict through \emph{role-decoupled conditioning}. We introduce five learned role tokens---country, region, city, latitude, and longitude---grouped into contextual and spatial sets, and apply lightweight semantic supervision to shape them. A separate projection pools the two groups and maps their concatenation to the single condition vector the frozen head expects, so the semantic supervision is decoupled from the final condition interface.
 
This makes \method\ a decode-side contribution that complements rather than competes with reasoning methods: better reasoning improves the semantic inputs, while \method\ improves how those semantics become a coordinate. Decoding through a frozen, geometry-aware head also clarifies where accuracy is won. Given an oracle semantic condition, the same frozen head localizes far more precisely, so the limiting factor is the condition supplied to the head, not the head itself. And because the decode is image-grounded and continuous rather than a discrete name lookup, an incorrect place name no longer forces a jump to the wrong city center: \method\ stays modestly closer to the truth than discrete geocoding even when the predicted semantics are imperfect. Our contributions are:
\begin{itemize}[leftmargin=1.2em,itemsep=0.2em]
    \item We identify a \emph{role conflict} in MLLM-conditioned generative geolocalization---discrete semantic supervision corrupts the continuous condition a frozen flow head needs---and propose \emph{role-decoupled conditioning} to resolve it.
    \item We instantiate it as \method: five role tokens with semantic supervision and contextual/spatial pooling, driving a frozen RFM spherical head through a trainable connector.
    \item On IM2GPS3K~\cite{vo2017revisiting}, \method\ improves over place-name geocoding (GLOBE~\cite{globe}) and reasoning-augmented prediction (GRE~\cite{gre}) at the $25/200/750$\,km thresholds, and remains competitive on an in-domain MP16 test.
\end{itemize}
\section{Related Work}

\paragraph{Worldwide image geolocalization.}
Classic worldwide geolocalization estimates a coordinate by retrieving visually similar geotagged images~\cite{TransGeo,cross_view_loc,zhu2021vigorcrossviewimagegeolocalization,zhang2023cross,vivanco2023geoclip} or by partitioning the Earth into geographic cells~\cite{weyand2016planet,seo2018,pramanick2022,haas2024pigeon}. Retrieval can preserve coordinate-level outputs when the reference set contains a similar view, whereas classification scales well but inherits discretization and boundary artifacts. Subsequent methods strengthen the visual side of this formulation with stronger encoders, geographic hierarchies, semantic segmentation, and scene context~\cite{pramanick2022translocator,clark2023geodecoder}. More recently, generative methods such as LocDiff~\cite{locdiff} and PLONK~\cite{around_the_world_in_80_timesteps} move beyond fixed grids and galleries by modeling geolocation as conditional coordinate generation. In parallel, MLLM-based methods use foundation models to infer location from explicit geographic cues, motivating a separate line of work discussed next.

\begin{figure}[t]
  \centering
  \includegraphics[width=\linewidth]{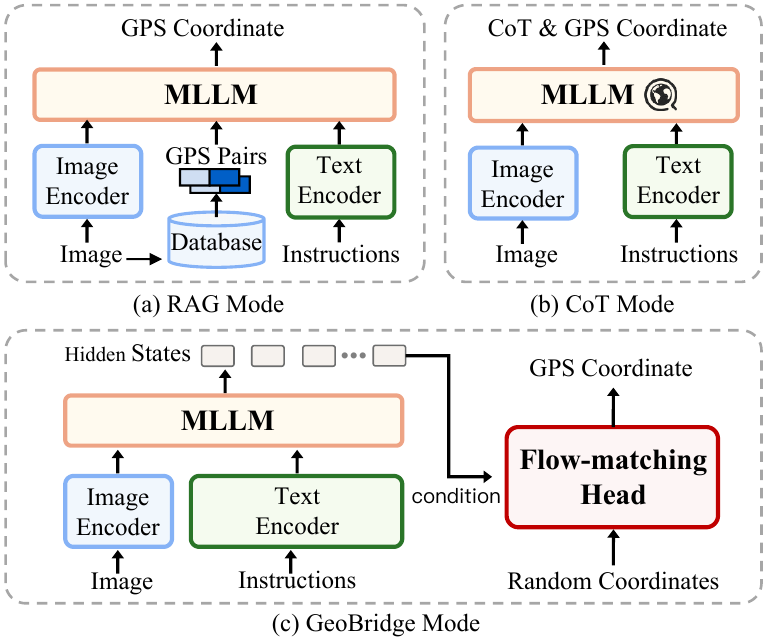}
  \caption{\textbf{Three modes of MLLM-based geolocation.}
  \emph{(a) RAG}: the MLLM is augmented with retrieved geotagged neighbors and predicts coordinates as text.
  \emph{(b) CoT}: the MLLM reasons over visual cues and emits a chain of thought together with coordinates.
  \emph{(c) \method}: instead of emitting coordinates as text, we read the MLLM's hidden states as a condition for a frozen flow-matching head that generates coordinates, decoupling semantic reasoning from continuous coordinate decoding.}
  \label{fig:modes}
\end{figure}

\paragraph{MLLM-based geolocation.}
Multimodal LLMs exploit explicit semantic cues---text, landmarks, road infrastructure, and regional visual style---and have produced two dominant strategies (Fig.~\ref{fig:modes}a,b). 
\emph{Retrieval-augmented} methods condition the MLLM on geotagged neighbors: Img2Loc~\cite{Img2Loc} and G3~\cite{G3} reframe geolocation as training-free generation over CLIP-retrieved candidates, GeoRanker~\cite{GeoRanker} introduces distance-aware ranking of retrieved candidates, and GeoToken~\cite{GeoToken} further combines geo-aligned retrieval with autoregressive S2-token decoding. 
\emph{Reasoning-based} methods instead strengthen intrinsic inference: GeoReasoner~\cite{GeoReasoner} fine-tunes on human geo-game reasoning traces, GaGA~\cite{gaga} introduces dynamic multi-round interaction into the geolocation task, GLOBE~\cite{globe} and GRE~\cite{gre} optimize visual-cue chains of thought, and GeoAgent~\cite{geoagent} aligns geolocation reasoning with geographic characteristics through expert-annotated CoT data and geo-aware reinforcement rewards. 
\method\ pursues a third mode (Fig.~\ref{fig:modes}c): it reads the MLLM's hidden states as a continuous condition for a frozen Riemannian flow head on $\mathbb{S}^{2}$, separating semantic reasoning from coordinate decoding without retrieving candidates or predicting discrete location tokens.

\paragraph{Learned interfaces between foundation models and decoders.}
Learnable query tokens and connector modules are widely used to bridge pretrained foundation-model representations and task-specific decoders. BLIP-2~\cite{li2023blip2} uses query tokens to connect frozen visual encoders with language models, while MetaQuery~\cite{metaquery} studies transferable query-based interfaces across modalities. Similar interface designs have also been used to adapt foundation-model hidden states to dense prediction tasks such as detection~\cite{kamath2021mdetrmodulateddetection,li2022groundedlanguageimagepretraining,chen2023shikraunleashingmultimodalllms,liu2024groundingdinomarryingdino} and segmentation~\cite{anchorseg,lisa,zou2022generalizeddecodingpixelimage,ren2024pixellmpixelreasoninglarge,rasheed2024glammpixelgroundinglarge}. GeoBridge follows this general connector paradigm, but the interface is especially delicate in our setting because the decoder is a continuous spherical flow head: the condition is not merely a class token or a natural-language prompt, but a latent that modulates a vector field on $S^2$.
\section{Method}
\label{sec:method}

\begin{figure*}[t]
  \centering
  \includegraphics[width=\linewidth]{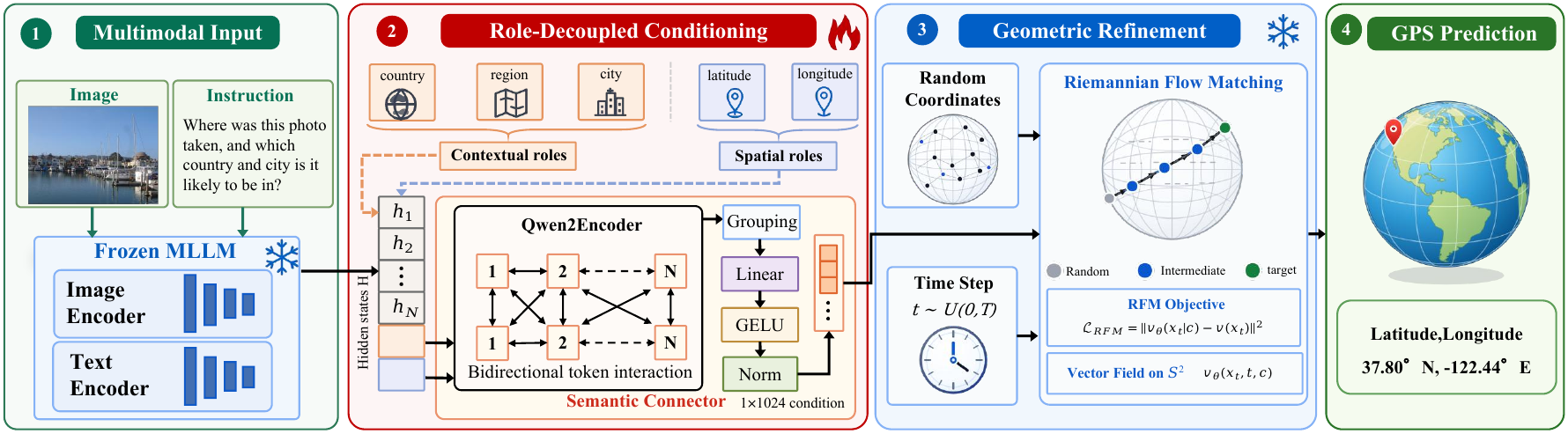}
  \caption{\textbf{\method\ architecture.}
  \textbf{(1)} A frozen MLLM encodes the image and a geographic instruction.
  \textbf{(2)} Five learned role tokens---three \emph{contextual} (country, region, city) and two \emph{spatial} (latitude, longitude)---are read from the MLLM and passed through a lightweight connector. Their contextual and spatial groups are pooled separately and projected into a single condition $\cond$, preserving the frozen head's single-token contract.
  \textbf{(3)} A frozen Riemannian flow-matching head consumes $\cond$ and transports a random point to a coordinate on the sphere $\Sph$.
  \textbf{(4)} The sampled point is the predicted GPS location. }
  \label{fig:architecture}
\end{figure*}

We present \method, a role-decoupled conditioning mechanism that connects a frozen semantic MLLM to a frozen spherical generative geolocation head (Fig.~\ref{fig:architecture}). \method\ is not a new coordinate generator: the RFM head is kept fixed. The contribution is the trainable interface that turns MLLM hidden states into the single continuous condition that head expects, and the way it injects semantic supervision without corrupting that condition.

\subsection{Problem Formulation}
\label{subsec:formulation}

Given an image $\img$, the goal is to predict a coordinate $\hat{\coord}\in\Sph$, scored by geodesic distance $\dist(\hat{\coord},\coord)$ to the ground truth. We view the system as a modular conditional generator:
\begin{equation}
    \hat{\coord} \sim g_{\phi}(\cdot \mid \cond), \qquad
    \cond = h_{\theta}\!\left(f_{\mathrm{MLLM}}(\img)\right),
    \label{eq:modular_predictor}
\end{equation}
where $f_{\mathrm{MLLM}}$ is a frozen MLLM, $g_{\phi}$ is a frozen RFM head pretrained to consume a single condition token, and $h_{\theta}$ is the \emph{only} trainable bridge. The bridge must satisfy two requirements at once: expose semantically meaningful geographic information from the MLLM, and preserve the continuous condition geometry the frozen spherical decoder was trained on.

\subsection{Role-Decoupled Conditioning}
\label{subsec:method_core}

\paragraph{Role conflict.}
Under naive supervision, these two requirements collide. Country and city labels are discrete and class-discriminative, encouraging decision boundaries between administrative categories; the RFM condition, by contrast, must vary smoothly with the coordinate posterior on $\Sph$. If a single vector must be both a country/city classifier state and the final flow condition, cross-entropy gradients improve semantic separability while distorting the geometry $g_{\phi}$ consumes. \method\ avoids this by decoupling the two roles across three components: role-structured tokens, role-specific semantic heads, and a projection buffer that restores the single-token condition the frozen head expects.

\paragraph{Role tokens.}
\method\ appends a total of five learned role tokens $\mathcal{R} = \mathcal{R}_{\mathrm{adm}} \cup \mathcal{R}_{\mathrm{coord}}$ after the image and geographic prompt, where
\begin{equation}
    \mathcal{R}_{\mathrm{adm}} = \{r_{\mathrm{cty}}, r_{\mathrm{reg}}, r_{\mathrm{city}}\}, \quad
    \mathcal{R}_{\mathrm{coord}} = \{r_{\mathrm{lat}}, r_{\mathrm{lon}}\},
    \label{eq:role_sets}
\end{equation}
for country, region, city, latitude, and longitude, with last-layer hidden states $\tokens\in\mathbb{R}^{5\times d_{\mathrm{m}}}$. The first three are \emph{contextual} roles (administrative/semantic prior); the last two are \emph{spatial} roles, giving coordinate-oriented information a place to enter the condition without sharing a token with country/city supervision. This split is purely architectural and does not require the MLLM to emit coordinates through these tokens. The MLLM is frozen; only the role-token embeddings and connector-side modules are trained.


\paragraph{Role-specific semantic heads.}
To inject multi-level geographic priors, the representations of the administrative roles $\mathcal{R}_{\mathrm{adm}}$ are regularized via auxiliary semantic classification tasks. Formally, the semantic prediction $\ell_r$ for a targeted role $r \in \mathcal{R}_{\mathrm{adm}}$ is computed as:
\begin{equation}
    \ell_r = A_r \, \mathrm{Norm}_r(\mathbf{h}_r) + b_r,
\end{equation}
where $\mathbf{h}_r$ denotes the corresponding hidden state. These auxiliary heads are optimized via cross-entropy losses against the ground-truth administrative labels. The overall geographic semantic loss is formulated as a joint optimization across the administrative granularities:
\begin{equation}
    \mathcal{L}_{\mathrm{sem}} = \sum_{r \in \mathcal{R}_{\mathrm{adm}}} \lambda_r \mathcal{L}_r^{\mathrm{CE}},
    \label{eq:semantic_loss}
\end{equation}
where $\lambda_r$ is the balancing coefficient for role $r$. Crucially, these auxiliary heads serve solely as regularizers to steer the hidden states toward interpretable geographic concepts, while the downstream projection buffer aggregates the refined representations for the final prediction.

\paragraph{Projection buffer and the single-token contract.}
A lightweight bidirectional transformer encoder $E_{\theta}$---a single Qwen2-style block---lets the role tokens exchange information, $\bar{\tokens}=E_{\theta}(W_{\mathrm{in}}\tokens)\in\mathbb{R}^{5\times d_{\mathrm{h}}}$. We then pool the contextual and spatial groups separately,
\begin{equation}
    \begin{aligned}
    \cond_{\mathrm{ctx}} &=
    \tfrac{1}{3}\!\left(\bar{\tokens}_{\mathrm{cty}}+\bar{\tokens}_{\mathrm{reg}}+\bar{\tokens}_{\mathrm{city}}\right), \\
    \cond_{\mathrm{spa}} &=
    \tfrac{1}{2}\!\left(\bar{\tokens}_{\mathrm{lat}}+\bar{\tokens}_{\mathrm{lon}}\right),
    \end{aligned}
    \label{eq:pooling}
\end{equation}
and map their concatenation to a single condition token,
\begin{equation}
    \cond =
    \mathrm{Norm}\!\left(\mathrm{GELU}\!\left(
    W_{\mathrm{out}}[\cond_{\mathrm{ctx}};\cond_{\mathrm{spa}}]\right)\right)
    \in \mathbb{R}^{1\times 1024}.
    \label{eq:projection_buffer}
\end{equation}
The frozen head was pretrained under a single-token condition contract; feeding it a sequence of role tokens would change its input distribution and force the frozen decoder onto an interface it never saw. Eq.~\eqref{eq:projection_buffer} keeps role structure upstream while preserving that contract exactly. Because $\cond$ still derives from the supervised encodings, supervision and conditioning are decoupled at the head interface rather than made fully independent---a distinction we test directly in our ablations.

\subsection{Training Objective}
\label{subsec:objective}

The frozen head defines a Riemannian flow-matching objective on the sphere~\cite{chen2024flow}. For an intermediate state $z_t\in\Sph$ at flow time $t$ with target tangent velocity $u_t$, the head predicts $v_{\phi}(z_t,t,\cond)$ and
\begin{equation}
    \mathcal{L}_{\mathrm{RFM}} =
    \mathbb{E}_{t,z_t}\!\left[
    \left\| v_{\phi}(z_t,t,\cond) - u_t \right\|^2_{T_{z_t}\Sph}\right].
    \label{eq:rfm_loss}
\end{equation}
With $\phi$ fixed, Eq.~\eqref{eq:rfm_loss} backpropagates only into the connector through $\cond$. Since the MLLM forward dominates training cost while the frozen head is cheap, we estimate Eq.~\eqref{eq:rfm_loss} with a \emph{1-to-N} expansion: for each image we reuse its single condition $\cond$ across $N{=}8$ independent flow samples, reducing gradient variance at no extra MLLM cost (derivation in the appendix). The full objective is
\begin{equation}
    \mathcal{L} = \widehat{\mathcal{L}}_{\mathrm{RFM}}^{(N)} + \mathcal{L}_{\mathrm{sem}},
    \label{eq:total_loss}
\end{equation}
which tunes only the role-token embeddings, connector encoder, projection buffer, and semantic heads. Keeping both the MLLM and the coordinate generator frozen is deliberate: any gain must come from the conditioning interface, not from retraining the backbone or enlarging the decoder.

\subsection{Training and Evaluation Policy}
\label{subsec:train_eval}

\begin{figure}[t]
  \centering
  \includegraphics[width=\linewidth]{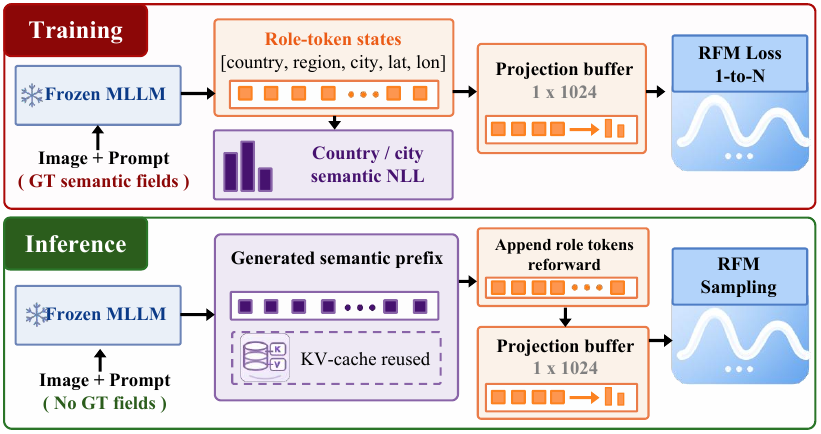}
  \caption{\textbf{Training and evaluation policy.} During training, ground-truth semantic fields are injected, whereas during evaluation they are unavailable.
  }
  \label{fig:train_eval_policy}
\end{figure}

The two passes differ only in how the semantic prefix is obtained (Fig.~\ref{fig:train_eval_policy}). \emph{At training}, ground-truth semantic text is injected as a teacher-forced prefix before the role tokens are read out, giving the connector a clean condition-learning signal; the semantic heads are evaluated here. \emph{At evaluation}, no ground truth is available: the frozen MLLM first generates the structured geographic text, then the role tokens are appended and reforwarded---reusing the prefix KV-cache---to produce the condition, with no semantic CE. All main results use this deployable generated-prefix path; the oracle (ground-truth-condition) setting is reported only as an upper bound in analysis.

\paragraph{Coordinate sampling.}
Given the condition $\cond$, the frozen head produces a coordinate by integrating its learned flow on the sphere, following~\cite{around_the_world_in_80_timesteps}. We draw a base sample $z_0$ from the prior on $\Sph$ (uniform over the Earth's surface) and solve the flow ODE $\dot{z}_t = v_{\phi}(z_t,t,\cond)$ from $t{=}0$ to $t{=}1$. Each step remains on the manifold through the exponential map,
\begin{equation}
    z_{t+\Delta t} = \exp_{z_t}\!\big(\Delta t\, v_{\phi}(z_t,t,\cond)\big),
    \qquad \Delta t = 1/K,
    \label{eq:rfm_sampling}
\end{equation}
using $K{=}32$ Euler steps. We take a single trajectory, so the endpoint is the prediction, $\hat{\coord}=z_1$. 
\section{Experiments}
\label{sec:experiments}

\subsection{Experimental Setup}
\label{subsec:setup}

\paragraph{Benchmarks.}
We evaluate on two worldwide image geolocalization benchmarks of web-sourced photographs: IM2GPS3K~\cite{vo2017revisiting} ($2{,}997$ images), the standard cross-domain test, and MP16-Reason-Test~\cite{globe} ($12{,}000$ images), an in-domain test drawn from the MP16-Pro distribution. 
Every method is scored against the same ground-truth split. 

\paragraph{Metrics.}
Distances are computed with the Haversine (great-circle) formula. We report the hit rate at 25, 200, 750, and 2500\,km, where @$k$ is the percentage of predictions within $k$\,km of the ground truth. The 25/200/750\,km thresholds are precision-relevant for city-, region-, and country-scale localization; @2500 is coarse and often saturated among modern MLLM baselines.

\paragraph{Baselines.}
We compare across geolocation paradigms; locally evaluated rows are starred, and per-method citations are given in the tables. Classification and retrieval methods predict a geographic cell or retrieve geotagged neighbors ~\cite{weyand2016planet,seo2018cplanet,ISNs,pramanick2022translocator,GeoDecoder}; embedding-alignment methods match images to a continuous location space~\cite{vivanco2023geoclip}; reasoning MLLMs infer a location through explicit visual-cue reasoning~\cite{GeoReasoner,gre,geoagent,gaea,globe}; place-name pipelines predict a place name and convert it to coordinates through a geocoding API~\cite{GeoReasoner,globe,gaea}, for which we run the released models under the original conversion (Microsoft Azure Map for \cite{globe} and Nominatim for \cite{gaea}); and generative methods sample coordinates from a flow head~\cite{around_the_world_in_80_timesteps}. \method\ is a decode-side mechanism, not a specialized retrieval/geocell system, and we make no leaderboard claim. We exclude the concurrent system ~\cite{geoagent,geo-r}, which strengthens upstream geographic reasoning rather than the decode and is therefore orthogonal to ours; a controlled comparison is also impractical (undisclosed training data, API-dependent geocoding).

\paragraph{Implementation.}
We train \method\ for one epoch on a 1M-image subset of MP16-Pro~\cite{G3} that is disjoint from the evaluation benchmarks, using learning rate $1\times10^{-4}$ and a cosine schedule. \method\ employs GLOBE~\cite{globe} as the base MLLM, utilizing five role tokens. The states of these tokens are encoded by the connector, then pooled into contextual and spatial groups, concatenated, and finally projected to a single $1024$-dimensional condition that is compatible with the frozen Riemannian-flow head~\cite{around_the_world_in_80_timesteps}. The MLLM backbone and RFM head are frozen; trainable parameters are limited to the new role-token embeddings, the connector modules, and the semantic auxiliary heads. The RFM loss uses 1-to-$N$ condition expansion with $N=8$; we derive the estimator in Appendix~\ref{app:one_to_n} and sweep $N$ in Appendix~\ref{app:n_sweep}. Further training details are provided in Appendix~\ref{app:training_details}.

\subsection{Main Results}
\label{subsec:main_results}
Table~\ref{tab:main_results} compares \method\ with standard geolocation references and recent MLLM-based systems on the two benchmarks; locally evaluated rows are starred.

\begin{table*}[t]
\centering
\caption{\textbf{Benchmark comparisons.} Accuracy (\%) at geodesic distance thresholds (km). $\star$ marks rows evaluated locally under our pipeline; for place-name methods we follow the original place-name-to-coordinate API conversion. Best in \textbf{bold}, second \underline{underlined}.}
\label{tab:main_results}
\begin{subtable}[t]{0.49\textwidth}
\centering
\caption{IM2GPS3K.}
\label{tab:main_im2gps3k}
\footnotesize
\setlength{\tabcolsep}{2.6pt}
\begin{tabular}{llcccc}
\toprule
Method & Source & @25 & @200 & @750 & @2500 \\
\midrule
PlaNet~\cite{weyand2016planet}      & ECCV'16    & 24.80 & 34.30 & 48.40 & 64.60 \\
ISNs~\cite{ISNs}        & ECCV'18    & 28.00 & 36.60 & 49.70 & 66.00 \\
Translocator~\cite{pramanick2022translocator}& ECCV'22    & 31.10 & 46.70 & 58.90 & 80.10 \\
GeoDecoder~\cite{clark2023geodecoder}  & CVPR'23    & 33.50 & 45.90 & 61.00 & 76.10 \\
GeoCLIP~\cite{vivanco2023geoclip}     & NeurIPS'23 & 34.47 & 50.65 & 69.67 & 83.82 \\
GRE~\cite{gre}         & NeurIPS'25 & 35.30  & 51.70  & 69.30  & \textbf{85.70} \\
\rowcolor{baselinegray}
GeoReasoner~\cite{GeoReasoner} & ICML'24    & 26.94 & 36.63 & 52.27 & 65.39 \\
\rowcolor{baselinegray}
GLOBE$\star$~\cite{globe}& NeurIPS'25 & \underline{36.95} & \underline{51.99} & \underline{69.88} & 83.99 \\
\rowcolor{baselinegray}
GAEA$\star$~\cite{gaea} & WACV'26    & 35.40 & 51.92 & 69.77 & 83.95 \\
\specialrule{\lightrulewidth}{0pt}{0pt}
\rowcolor{methodyellow}
\textbf{\method{}}$\star$ & This work & \textbf{38.67} & \textbf{52.89} & \textbf{70.37} & \underline{84.42} \\
\specialrule{\heavyrulewidth}{0pt}{0pt}
\end{tabular}
\end{subtable}
\hfill
\begin{subtable}[t]{0.49\textwidth}
\centering
\caption{MP16-Reason-Test.}
\label{tab:mp16_reason_test}
\footnotesize
\setlength{\tabcolsep}{2.6pt}
\begin{tabular}{llcccc}
\toprule
Method & Source & @25 & @200 & @750 & @2500 \\
\midrule
ISNs~\cite{ISNs}        & ECCV'18    & 47.38 & 55.88 & 68.48 & 80.92 \\
GeoCLIP~\cite{vivanco2023geoclip} & NeurIPS'23 & 52.52 & 66.85 & 84.07 & \underline{93.33} \\
Hybrid~\cite{OSV5M}      & CVPR'24    & 16.53 & 28.72 & 50.31 & 71.47 \\
RFM-YFCC~\cite{around_the_world_in_80_timesteps}    & CVPR'25    & 46.64 & 60.46 & 77.97 & 91.96 \\
GRE~\cite{gre}         & NeurIPS'25 & - & - & - & - \\
\rowcolor{baselinegray}
GeoReasoner~\cite{GeoReasoner} & ICML'24    & 40.44 & 50.91 & 68.01 & 79.68 \\
\rowcolor{baselinegray}
GLOBE$\star$~\cite{globe} & NeurIPS'25 & \textbf{60.95} & \textbf{74.00} & \underline{86.12} & 92.98 \\
\rowcolor{baselinegray}
GAEA$\star$~\cite{gaea} & WACV'26    & 26.13 & 41.48 & 65.60 & 82.84 \\
\specialrule{\lightrulewidth}{0pt}{0pt}
\rowcolor{methodyellow}
\textbf{\method{}}$\star$ & This work & \underline{57.44} & \underline{72.39} & \textbf{87.08} & \textbf{94.20} \\
\specialrule{\heavyrulewidth}{0pt}{0pt}
\end{tabular}
\end{subtable}
\end{table*}

On IM2GPS3K (Table~\ref{tab:main_im2gps3k}), among the non-retrieval MLLM decoding baselines in Table~\ref{tab:main_results}, \method\ attains the strongest accuracy at 25, 200, and 750\,km, and trails only GRE at the coarse, near-saturated 2500\,km band. It improves over both place-name-to-API baselines (GeoReasoner, GLOBE, GAEA) and the reasoning-augmented GRE at fine scale. This supports an algorithmic point rather than a leaderboard one: place-name geocoding discards information by committing to a discrete name and direct-coordinate MLLMs emit a single point through the language head, whereas \method\ keeps a generative spherical decoder and forms its condition from MLLM semantics through a role-decoupled interface, improving fine-scale accuracy without modifying the frozen head.
 
\textbf{Cross-benchmark divergence.} On MP16-Reason-Test (Table~\ref{tab:mp16_reason_test}) \method\ overtakes place-name geocoding (GLOBE) at 750 and 2500\,km, and GLOBE leads only at the two finest thresholds, by a small margin. This fine-threshold gap reflects how the benchmark defines ground truth, not a localization weakness. MP16-Reason-Test coordinates are city-center-aligned, so once the city is named correctly, geocoding to the city center lands almost exactly on the target: among GLOBE's predictions, the subset with both country and city correct ($n{=}5815$) reaches $98.64/99.57/99.67/99.93$ at @25/@200/@750/@2500. On the same subset, \method\ is precise within the city but spreads continuously around the true area rather than snapping to the city-center point the metric treats as ground truth. This is why GLOBE's edge is confined to 25/200\,km, where landing on the center is decisive, and reverses at 750/2500\,km, where center-snapping no longer helps. The city-center bonus, however, only applies when the city is named correctly, and that coverage is domain-dependent: GLOBE---also \method's semantic backbone---is both-country-and-city-correct on $5815/12000$ ($48\%$) of MP16-Reason-Test but only $736/2997$ ($25\%$) of cross-domain IM2GPS3K. With the discrete bonus covering far fewer images there, \method's continuous estimate prevails on the larger remainder and leads at the precision thresholds. Which decode paradigm wins thus depends on how often the discrete city is correct, not on localization quality alone; we report both benchmarks to make this dependence explicit.

\subsection{Analysis}
\label{subsec:upper_bound}
All analysis in this section is on IM2GPS3K.
 
\textbf{The condition, not the head, is the bottleneck.} To locate the limiting factor, we replace the deployable condition with a teacher-forced \emph{oracle} role condition built from ground-truth country/region/city text (Table~\ref{tab:oracle_gap}). The oracle row is not a deployable accuracy claim; it isolates whether the frozen head and five-role connector can exploit a correct semantic--spatial condition. Given that condition, @25 rises from $38.67$ to $71.67$ and the median error falls from $145.77$ to $11.67$\,km---the frozen RFM head is far from saturated. The deployable limit is therefore the quality of the condition supplied by the MLLM, not the capacity of the coordinate head. This validates the premise behind \method: rather than strengthen the coordinate head, we keep a strong frozen decoder and learn a better, image-grounded condition for it.
 
\begin{table}[t]
\centering
\caption{\textbf{Condition ceiling for the frozen RFM head.} The oracle row uses teacher-forced role conditions and is not a deployable result; it measures whether the frozen head and five-role connector can exploit a high-quality condition.}
\label{tab:oracle_gap}
\small
\setlength{\tabcolsep}{3.6pt}
\begin{tabular}{lccccc}
\toprule
Condition & @25 & @200 & @750 & @2500 & Median \\
\midrule
Deployable & 38.67 & 52.89 & 70.37 & 84.42 & 145.77 \\
Oracle     & \textbf{71.67} & \textbf{88.51} & \textbf{95.50} & \textbf{99.19} & \textbf{11.67} \\
\bottomrule
\end{tabular}
\end{table}

\begin{figure*}[t]
  \centering
  \includegraphics[width=\textwidth]{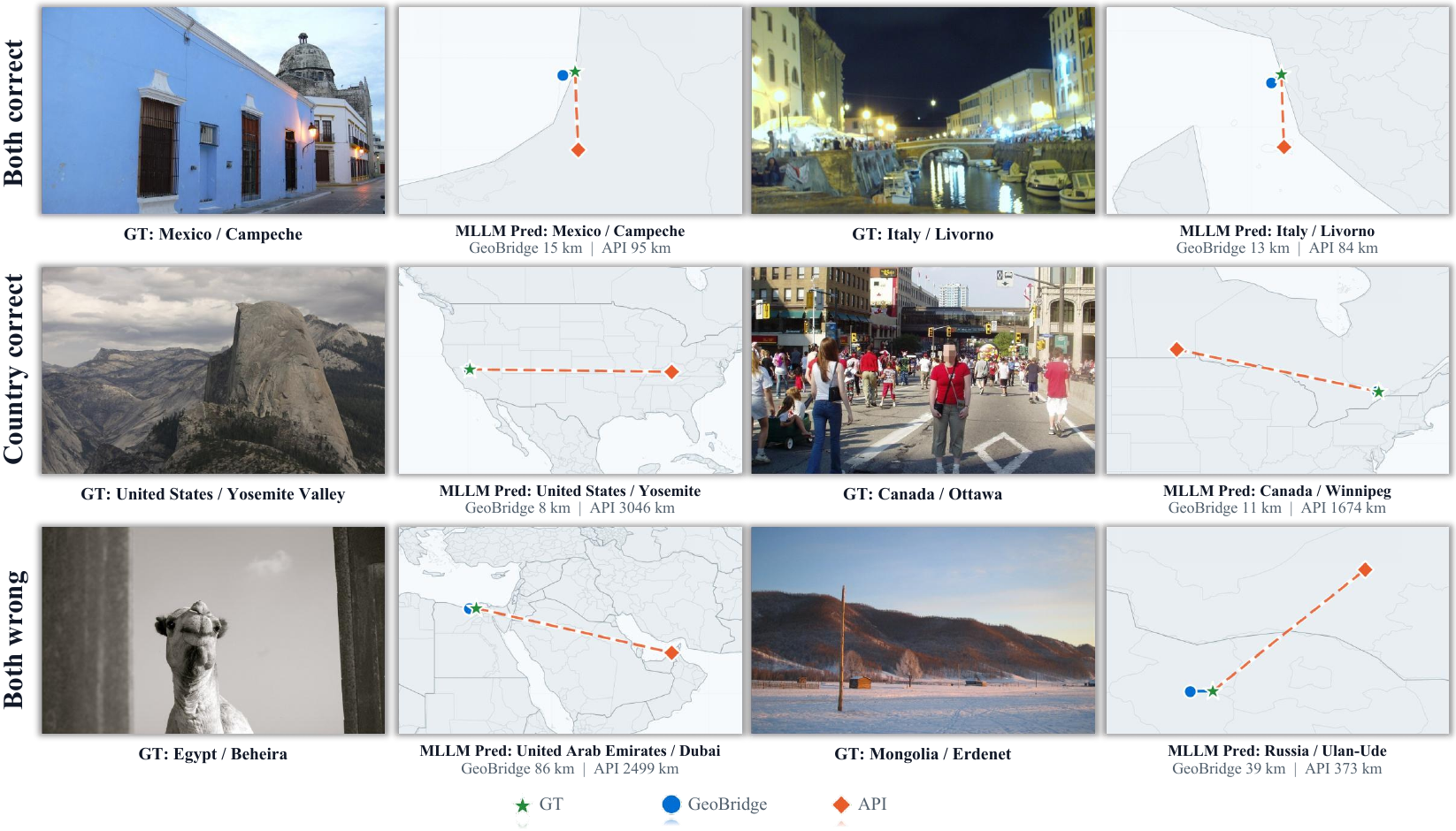}
  \caption{\textbf{Robustness to imperfect semantics (IM2GPS3K).} Two examples per bucket; each map shows the ground truth (green star), \method\ (blue circle), and place-name geocoding (orange diamond). }
  \label{fig:qualitative_cases}
\end{figure*}

\textbf{An image-grounded decoder degrades gracefully.} The two decode paradigms fail very differently when the semantics are imperfect, and this is where \method's design pays off. A place-name pipeline sees only the discrete predicted name at decode time, so a wrong city name geocodes to the wrong city center---an error as large as the naming mistake. \method\ instead conditions on the image-grounded MLLM state and decodes continuously, so a semantic error is softened by the visual evidence rather than amplified. Grouping IM2GPS3K predictions by the correctness of the generated country/city condition (Table~\ref{tab:semantic_buckets}), \method\ attains lower mean error than place-name geocoding in every regime. When both labels are correct its continuous estimate localizes within the city rather than snapping to the center ($42.9$ vs.\ $57.3$\,km); and when the city or even the country is wrong, conditioning on the image-grounded state keeps it at least as accurate as geocoding ($441.5$ vs.\ $463.5$\,km; $4263.5$ vs.\ $4384.4$\,km) instead of being dragged to a wrong administrative center. \method\ thus never trades away accuracy for conditioning on the MLLM, and Fig.~\ref{fig:qualitative_cases} shows representative cases. Only $736$ of the $2{,}997$ images fall in the both-correct bucket, so most residual error is upstream---\method\ improves and stabilizes the semantics-to-coordinate decode but does not repair the reasoning itself, making it complementary to stronger reasoning or CoT-conditioned backbones that would improve the condition it decodes.
 
\begin{table}[t]
\centering
\caption{\textbf{Robustness across semantic regimes (IM2GPS3K).} Mean Haversine error (km) per bucket, grouped by whether the generated condition names the correct country and city. \method\ attains lower mean error than place-name geocoding in every bucket, including when the discrete semantics are wrong.}
\label{tab:semantic_buckets}
\small
\setlength{\tabcolsep}{4pt}
\begin{tabular}{lrrr}
\toprule
Bucket & Count & \method & Geocoding \\
\midrule
Country + city correct      & 736  & \textbf{42.92}   & 57.25 \\
Country correct, city wrong & 1355 & \textbf{441.49}  & 463.51 \\
Country + city wrong        & 906  & \textbf{4263.54} & 4384.41 \\
\bottomrule
\end{tabular}
\end{table}
 
Fig.~\ref{fig:qualitative_cases} contrasts \method\ with place-name geocoding on two examples from each bucket. When the condition is correct both land near the ground truth; when only the country is correct, geocoding jumps to a wrong city center while \method\ stays within the right region; and when the condition is wrong, geocoding follows the wrong name across continents while \method's image-grounded estimate remains comparatively close. It is the visual evidence, not the discrete label, that keeps \method\ anchored.
 
\section{Ablations}
\label{sec:ablations}
Unless otherwise specified, all experiments in this section are conducted on IM2GPS3K.

\subsection{Component Ablation}
\label{subsec:ablation}

\begin{table}[t]
\centering
\caption{\textbf{Conditioning-interface ablation under the oracle condition.} The condition path is built up one design element at a time and evaluated on IM2GPS3K with the teacher-forced ground-truth semantic condition, so each variant's representational ceiling is measured in isolation from upstream semantic-prediction errors. We report Acc@$k$ (\%); best per column in \textbf{bold}. The last row is the Oracle entry of Table~\ref{tab:oracle_gap}.}
\label{tab:ablation}
\small
\setlength{\tabcolsep}{4pt}
\begin{tabular}{lcccc}
\toprule
Conditioning variant & @25 & @200 & @750 & @2500 \\
\midrule
Single-token, MLP                 & 38.61 & 63.86 & 84.48 & 94.09 \\
5 tokens, ungrouped, MLP          & 37.00 & 64.03 & 80.85 & 92.09 \\
5 tokens, grouped, MLP            & 42.38 & 69.87 & 84.42 & 95.70 \\
5 tokens, grouped, Q2Enc. & 61.06 & 80.07 & 93.18 & 98.69 \\
\textbf{Full \method{} (+ role CE)} & \textbf{71.67} & \textbf{88.51} & \textbf{95.50} & \textbf{99.19} \\
\bottomrule
\end{tabular}
\end{table}
 
Table~\ref{tab:ablation} builds \method's conditioning interface one design element at a time. To measure each variant's representational ceiling rather than its sensitivity to upstream errors, we evaluate under the \emph{oracle} condition---the teacher-forced ground-truth semantic text of Sec.~\ref{subsec:upper_bound}---so the final row coincides with the Oracle entry of Table~\ref{tab:oracle_gap}. All variants share the same training data, schedule, frozen MLLM backbone, and frozen RFM head; what varies is the condition path: the role-token set, the grouping of those tokens, the connector module, and the auxiliary role cross-entropy.
 
\textbf{Structure matters more than token count.} Replacing the single condition token with five \emph{ungrouped} role tokens does not help: @25 even dips ($38.6{\to}37.0$) and the coarser thresholds fall ($84.5{\to}80.9$ at @750), so adding latent tokens alone does not improve the condition. Grouping those role tokens into contextual and spatial sets instead yields a clear gain at every threshold ($37.0{\to}42.4$ at @25, $80.9{\to}84.4$ at @750), showing that structured semantic decomposition, not token budget, is the active ingredient.
 
\textbf{Connector capacity and role supervision close the gap.} Replacing the MLP connector with Q2Enc, a Qwen2-style transformer-block encoder, lifts the ceiling sharply ($42.4{\to}61.1$ at @25). Adding the auxiliary role cross-entropy---which, decoupled from the condition geometry by \method's projection buffer, stabilizes role specialization rather than corrupting it---completes \method\ and reaches the oracle ceiling ($71.67$/$88.51$/$95.50$/$99.19$). Oracle ceilings need not translate into deployable gains, however: Appendix~\ref{app:deployable_ablation} repeats this ablation under the deployable generated-condition setting, where the connector accounts for most of the realized improvement and the role cross-entropy is approximately neutral---consistent with deployable accuracy being gated by the generated semantic prefix.

\subsection{Coordinate Sampler}
\label{subsec:sampler}
The frozen head can be trained with Euclidean flow matching (FM) or Riemannian flow matching (RFM) on the sphere; the two differ only in the sampler construction and the flow loss, with all other settings fixed. We compare them at the \emph{oracle} condition, which probes the head in isolation from generated-condition quality (Table~\ref{tab:sampler_oracle}). RFM is at least as strong as FM at every threshold, with the largest gain at @25 ($71.67$ vs. $66.80$) and a tie at @750 ($95.50$ for both), supporting the geometry-aware decoder choice; we use RFM throughout.

\begin{table}[t]
\centering
\caption{\textbf{Oracle-condition upper bounds with different coordinate samplers.} Both rows use teacher-forced semantic conditions and are not deployable results. 
}
\label{tab:sampler_oracle}
\small
\setlength{\tabcolsep}{5.0pt}
\begin{tabular}{lcccc}
\toprule
Sampler/head & @25 & @200 & @750 & @2500 \\
\midrule
Flow Matching & 66.80 & 88.00 & 95.50 & 98.60 \\
Riemannian Flow Matching & \textbf{71.67} & \textbf{88.51} & 95.50 & \textbf{99.19} \\
\bottomrule
\end{tabular}
\end{table}

\subsection{Integration Steps at Inference}
\label{subsec:steps}
\begin{figure}[t]
  \centering
  \includegraphics[width=0.9\linewidth]{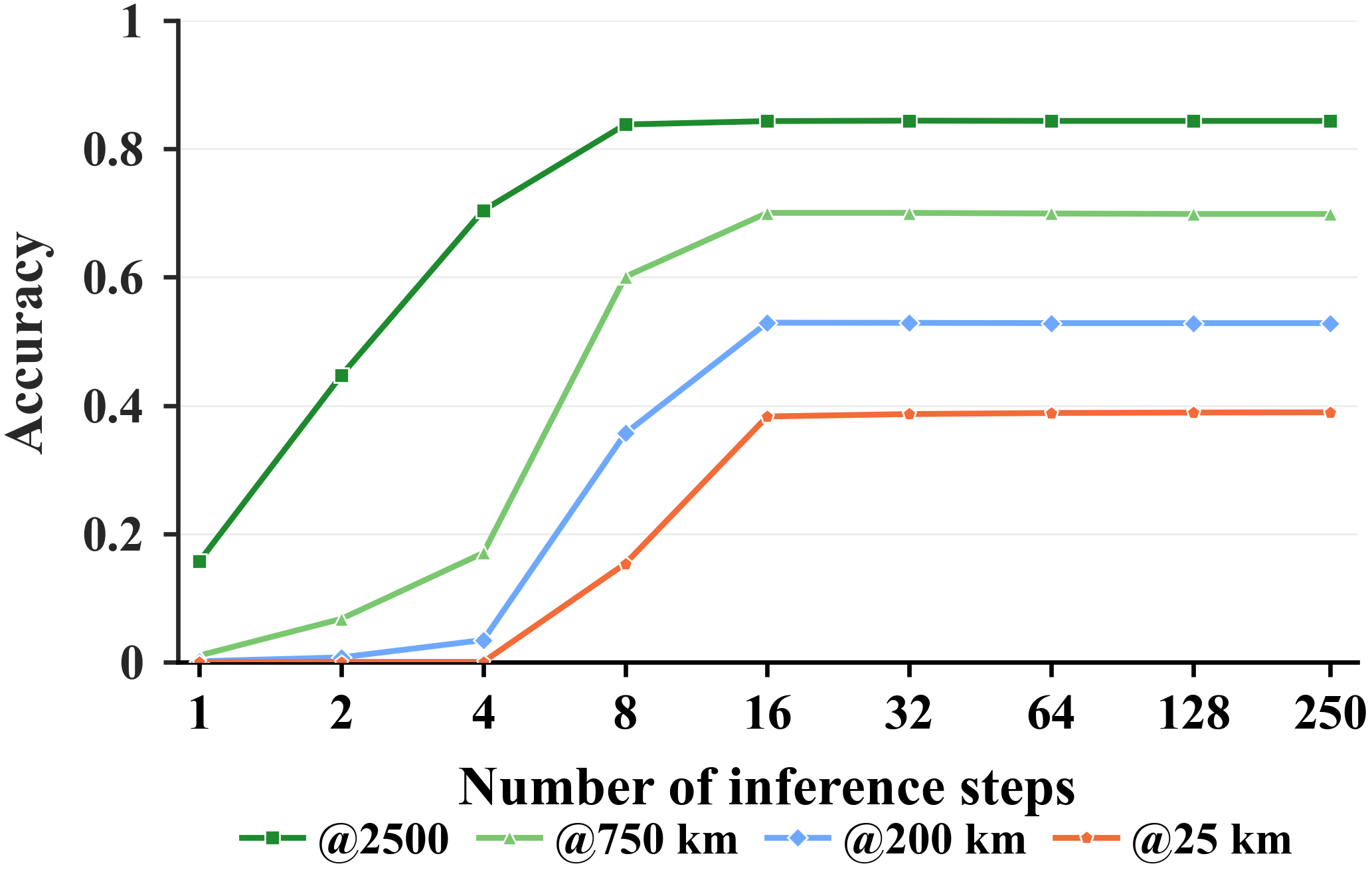}
  \caption{\textbf{Accuracy vs.\ number of inference steps.} All thresholds saturate by roughly $16$ Euler steps; our default of $32$ is well past the plateau.}
  \label{fig:steps_sweep}
\end{figure}

\method\ samples a coordinate by integrating the frozen flow for $K$ Euler steps (Sec.~\ref{subsec:train_eval}). Fig.~\ref{fig:steps_sweep} sweeps $K$ at test time with all trained components fixed. Accuracy rises steeply up to roughly $16$ steps and then plateaus across every threshold, so our default $K{=}32$ sits just past the plateau and can be reduced further for faster inference at negligible accuracy cost. The condition is computed once per image, so the only added cost is $K$ lightweight head evaluations.

\subsection{Backbone Transfer}
\label{subsec:backbone}
To probe how much the backbone shapes the condition, we swap the frozen MLLM under a deliberately minimal interface---a single condition token, RFM loss, and no semantic supervision---keeping everything else fixed (Table~\ref{tab:backbone}). These runs predate the five-role design and are therefore not comparable to full \method; they isolate the backbone's effect on condition quality. We evaluate here under the deployable condition rather than the oracle one: the three backbones train almost identically and diverge chiefly in the semantic prefix they predict at inference, so only the deployable setting---where each backbone supplies its own generated condition---surfaces that difference, whereas teacher-forcing a shared ground-truth condition would mask it. This mirrors the connector ablation's use of the oracle condition: each setting fixes the factor it is not testing. Stronger geographic backbones indeed yield better conditions: a supervised-fine-tuned Qwen2.5-VL and GLOBE both improve over the base model across most thresholds. This reinforces the upper-bound finding that deployable accuracy is governed by the semantic condition the backbone supplies, and shows the generative-decoder interface is not tied to a single MLLM.

\begin{table}[t]
\centering
\caption{\textbf{Single-token backbone diagnostics.} 
These runs replace the frozen semantic backbone while keeping a one-token condition interface. 
They predate the five-role role-decoupled design, so they are diagnostic transfer evidence rather than matched \method\ ablations.
}
\label{tab:backbone}
\small
\setlength{\tabcolsep}{5.0pt}
\begin{tabular}{lcccc}
\toprule
Frozen backbone & @25 & @200 & @750 & @2500 \\
\midrule
Qwen2.5-VL~\cite{Qwen2.5-VL} & 21.7 & 42.6 & 65.3 & 82.5 \\
Qwen2.5-VL SFT & \textbf{23.3} & 48.7 & 67.8 & 84.3 \\
GLOBE & 21.6 & \textbf{48.9} & \textbf{67.9} & \textbf{84.8} \\
\bottomrule
\end{tabular}
\end{table}

\section{Conclusion}

We presented \method, a role-decoupled conditioning scheme for MLLM-guided generative geolocalization. Supervised semantic role tokens and a separate context/spatial projection feed a frozen RFM head without altering its single-token condition contract, resolving the role conflict between discrete semantic supervision and continuous coordinate geometry. On IM2GPS3K, this decode-side mechanism improves fine-scale accuracy over the MLLM-based and place-name baselines we compare. Our analysis locates the remaining gap upstream---in the semantic condition the MLLM supplies, not the coordinate head---pointing to distribution-aligned condition learning and its composition with stronger reasoning as the natural next step.

\section*{Acknowledgements}
This work was supported in part by the Key Deployment Program of the Chinese Academy of Sciences, China under Grant KGFZD-145-25-39, the National Natural Science Foundation of China under Grants 62272438, and Beijing Natural Science Foundation L25700.

{
    \small
    \bibliographystyle{ieeenat_fullname}
    \bibliography{references}

@inproceedings{weyand2016planet,
  title = {PlaNet: Photo Geolocation with Convolutional Neural Networks},
  author = {Weyand, Tobias and Kostrikov, Ilya and Philbin, James},
  booktitle = {European Conference on Computer Vision},
  pages = {37--55},
  year = {2016}
}

@inproceedings{vo2017revisiting,
  title = {Revisiting IM2GPS in the Deep Learning Era},
  author = {Vo, Nam and Jacobs, Nathan and Hays, James},
  booktitle = {Proceedings of the IEEE International Conference on Computer Vision},
  pages = {2621--2630},
  year = {2017}
}

@inproceedings{seo2018cplanet,
  title = {CPlaNet: Enhancing Image Geolocalization by Combinatorial Partitioning of Maps},
  author = {Seo, Paul Hongsuck and Weyand, Tobias and Sim, Jack and Han, Bohyung},
  booktitle = {Proceedings of the European Conference on Computer Vision},
  pages = {536--551},
  year = {2018}
}

@inproceedings{pramanick2022translocator,
  title = {Where in the World Is This Image? Transformer-Based Geo-Localization in the Wild},
  author = {Pramanick, Shraman and Nowara, Ewa M. and Gleason, Joshua and Castillo, Carlos D. and Chellappa, Rama},
  booktitle = {European Conference on Computer Vision},
  pages = {196--215},
  year = {2022}
}

@misc{gaga,
      title={GaGA: Towards Interactive Global Geolocation Assistant}, 
      author={Zhiyang Dou and Zipeng Wang and Xumeng Han and Guorong Li and Zhipei Huang and Zhenjun Han},
      year={2025},
      eprint={2412.08907},
      archivePrefix={arXiv},
      primaryClass={cs.CV},
      url={https://arxiv.org/abs/2412.08907}, 
}

@article{geoagent,
  title={GeoAgent: Learning to Geolocate Everywhere with Reinforced Geographic Characteristics},
  author={Jin, Modi and Zhang, Yiming and Sun, Boyuan and Zhang, Dingwen and Cheng, Ming-Ming and Hou, Qibin},
  journal={arXiv preprint arXiv:2602.12617},
  year={2026}
}

@misc{geo-r,
      title={Vision-Language Reasoning for Geolocalization: A Reinforcement Learning Approach}, 
      author={Biao Wu and Meng Fang and Ling Chen and Ke Xu and Tao Cheng and Jun Wang},
      year={2026},
      eprint={2601.00388},
      archivePrefix={arXiv},
      primaryClass={cs.CL},
      url={https://arxiv.org/abs/2601.00388}, 
}

@inproceedings{clark2023geodecoder,
  title = {Where We Are and What We're Looking At: Query Based Worldwide Image Geo-Localization Using Hierarchies and Scenes},
  author = {Clark, Brandon and Kerrigan, Alec and Kulkarni, Parth P. and Cepeda, Vicente Vivanco and Shah, Mubarak},
  booktitle = {Proceedings of the IEEE/CVF Conference on Computer Vision and Pattern Recognition},
  year = {2023}
}

@inproceedings{vivanco2023geoclip,
  title = {GeoCLIP: CLIP-Inspired Alignment between Locations and Images for Effective Worldwide Geo-Localization},
  author = {Vivanco Cepeda, Vicente and Nayak, Gaurav Kumar and Shah, Mubarak},
  booktitle = {Advances in Neural Information Processing Systems},
  volume = {36},
  pages = {8690--8701},
  year = {2023}
}

@inproceedings{haas2024pigeon,
  title = {PIGEON: Predicting Image Geolocations},
  author = {Haas, Lukas and Skreta, Michal and Alberti, Silas and Finn, Chelsea},
  booktitle = {Proceedings of the IEEE/CVF Conference on Computer Vision and Pattern Recognition},
  pages = {12893--12902},
  year = {2024}
}

@string{ECCV = "Proceedings of the European Conference on Computer Vision"}

@string{CVPR = "IEEE Conference on Computer Vision and Pattern Recognition"}

@inproceedings{cross_view_loc,
  title={Cross-View Image Sequence Geo-localization},
  author={Zhang, Xiaohan and Sultani, Waqas and Wshah, Safwan},
  booktitle={WACV},
  year={2023}
}

@inproceedings{TransGeo,
  title={{TransGeo: Transformer Is All You Need for Cross-view Image Geo-localization}},
  author={Zhu, Sijie and Shah, Mubarak and Chen, Chen},
  booktitle=CVPR,
  pages={1162-1171},
  year={2022}
}

@inproceedings{seo2018,
  title={{CPlaNet: Enhancing Image Geolocalization by Combinatorial Partitioning of Maps}},
  author={Seo, Paul Hongsuck and Weyand, Tobias and Sim, Jack and Han, Bohyung},
  booktitle=ECCV,
  pages={536-551},
  year={2018}
}

@inproceedings{OSV5M,
  title={OpenStreetView-5M: The Many Roads to Global Visual Geolocation},
  author={Astruc, Guillaume and Dufour, Nicolas and Siglidis, Ioannis and Aronssohn, Constantin and Bouia, Nacim and Fu, Stephanie and Loiseau, Romain and Nguyen, Van Nguyen and Raude, Charles and Vincent, Elliot and others},
  booktitle={Proceedings of the IEEE/CVF Conference on Computer Vision and Pattern Recognition},
  pages={21967--21977},
  year={2024}
}

@inproceedings{CLIP,
  title={Learning transferable visual models from natural language supervision},
  author={Radford, Alec and Kim, Jong Wook and Hallacy, Chris and Ramesh, Aditya and Goh, Gabriel and Agarwal, Sandhini and Sastry, Girish and Askell, Amanda and Mishkin, Pamela and Clark, Jack and others},
  booktitle={International conference on machine learning},
  pages={8748--8763},
  year={2021},
  organization={PMLR}
}

@misc{GeoReasoner,
      title={GeoReasoner: Geo-localization with Reasoning in Street Views using a Large Vision-Language Model}, 
      author={Ling Li and Yu Ye and Bingchuan Jiang and Wei Zeng},
      booktitle={International Conference on Machine Learning (ICML)},
      year={2024}
}

@article{GeoRanker,
  title={GeoRanker: Distance-Aware Ranking for Worldwide Image Geolocalization},
  author={Jia, Pengyue and Park, Seongheon and Gao, Song and Zhao, Xiangyu and Li, Yixuan},
  journal={arXiv preprint arXiv:2505.13731},
  year={2025}
}

@inproceedings{globe,
  title={Recognition through Reasoning: Reinforcing Image Geo-localization with Large Vision-Language Models},
  author={Li, Ling and Zhou, Yao and Liang, Yuxuan and Tsung, Fugee and Wei, Jiaheng},
  booktitle={Advances in Neural Information Processing Systems},
  year={2025}
}

@inproceedings{Img2Loc,
  title={Img2Loc: Revisiting image geolocalization using multi-modality foundation models and image-based retrieval-augmented generation},
  author={Zhou, Zhongliang and Zhang, Jielu and Guan, Zihan and Hu, Mengxuan and Lao, Ni and Mu, Lan and Li, Sheng and Mai, Gengchen},
  booktitle={Proceedings of the 47th international acm sigir conference on research and development in information retrieval},
  pages={2749--2754},
  year={2024}
}

@article{G3,
  title={G3: an effective and adaptive framework for worldwide geolocalization using large multi-modality models},
  author={Jia, Pengyue and Liu, Yiding and Li, Xiaopeng and Zhao, Xiangyu and Wang, Yuhao and Du, Yantong and Han, Xiao and Wei, Xuetao and Wang, Shuaiqiang and Yin, Dawei},
  journal={Advances in Neural Information Processing Systems},
  volume={37},
  pages={53198--53221},
  year={2024}
}

@inproceedings{zhang2023cross,
  title={Cross-view geo-localization via learning disentangled geometric layout correspondence},
  author={Zhang, Xiaohan and Li, Xingyu and Sultani, Waqas and Zhou, Yi and Wshah, Safwan},
  booktitle={Proceedings of the AAAI Conference on Artificial Intelligence},
  volume={37},
  number={3},
  pages={3480--3488},
  year={2023}
}

@article{GeoDecoder,
  title={GeoDecoder: Empowering Multimodal Map Understanding},
  author={Qi, Feng and Dai, Mian and Zheng, Zixian and Wang, Chao},
  journal={arXiv preprint arXiv:2401.15118},
  year={2024}
}

@inproceedings{ISNs,
  title={Geolocation estimation of photos using a hierarchical model and scene classification},
  author={Muller-Budack, Eric and Pustu-Iren, Kader and Ewerth, Ralph},
  booktitle={Proceedings of the European conference on computer vision (ECCV)},
  pages={563--579},
  year={2018}
}

@article{shen_2017_streetvizor,
   author = {Shen, Qiaomu and Zeng, Wei and Ye, Yu and Mueller Arisona, Stefan and Schubiger, Simon and Burkhard, Remo and Qu, Huamin},
   title = {{StreetVizor: Visual Exploration of Human-Scale Urban Forms Based on Street Views}},
   journal = {IEEE Transactions on Visualization and Computer Graphics},
   volume = {24},
   number = {1},
   year = {2018},
   pages = {1004-1013}
}

@inproceedings{pramanick2022,
  title={{Where in the World is this Image? Transformer-based Geo-localization in the Wild}},
  author={Pramanick, Shraman and Nowara, Ewa M and Gleason, Joshua and Castillo, Carlos D and Chellappa, Rama},
  booktitle=ECCV,
  pages={196-215},
  year={2022}
}

@inproceedings{Im2gps,
  title={Im2gps: estimating geographic information from a single image},
  author={Hays, James and Efros, Alexei A},
  booktitle={CVPR},
  year={2008},
}

@inproceedings{around_the_world_in_80_timesteps,
  title={Around the world in 80 timesteps: A generative approach to global visual geolocation},
  author={Dufour, Nicolas and Kalogeiton, Vicky and Picard, David and Landrieu, Loic},
  booktitle={Proceedings of the Computer Vision and Pattern Recognition Conference},
  pages={23016--23026},
  year={2025}
}

@article{Qwen2.5-VL,
  title={Qwen2.5-VL Technical Report},
  author={Bai, Shuai and Chen, Keqin and Liu, Xuejing and Wang, Jialin and Ge, Wenbin and Song, Sibo and Dang, Kai and Wang, Peng and Wang, Shijie and Tang, Jun and Zhong, Humen and Zhu, Yuanzhi and Yang, Mingkun and Li, Zhaohai and Wan, Jianqiang and Wang, Pengfei and Ding, Wei and Fu, Zheren and Xu, Yiheng and Ye, Jiabo and Zhang, Xi and Xie, Tianbao and Cheng, Zesen and Zhang, Hang and Yang, Zhibo and Xu, Haiyang and Lin, Junyang},
  journal={arXiv preprint arXiv:2502.13923},
  year={2025}
}

@misc{gre,
  title={Gre suite: Geo-localization inference via fine-tuned vision-language models and enhanced reasoning chains},
  author={Wang, Chun and Ye, Xiaojun and Pan, Xiaoran and Pan, Zihao and Wang, Haofan and Song, Yiren},
  booktitle={Advances in Neural Information Processing Systems},
  year={2025}
}

@misc{gaea,
      title={GAEA: A Geolocation Aware Conversational Model}, 
      author={Ron Campos and Ashmal Vayani and Parth Parag Kulkarni and Rohit Gupta and Aritra Dutta and Mubarak Shah},
      year={2025},
      eprint={2503.16423},
      archivePrefix={arXiv},
      primaryClass={cs.CV},
      url={https://arxiv.org/abs/2503.16423}, 
}

@article{lipman2022flow,
  title={Flow matching for generative modeling},
  author={Lipman, Yaron and Chen, Ricky TQ and Ben-Hamu, Heli and Nickel, Maximilian and Le, Matt},
  journal={arXiv preprint arXiv:2210.02747},
  year={2022}
}

@inproceedings{chen2024flow,
  title={Flow matching on general geometries},
  author={Chen, Ricky TQ and Lipman, Yaron},
  booktitle={International Conference on Learning Representations},
  volume={2024},
  pages={47922--47945},
  year={2024}
}

@inproceedings{li2023blip2,
  title={Blip-2: Bootstrapping language-image pre-training with frozen image encoders and large language models},
  author={Li, Junnan and Li, Dongxu and Savarese, Silvio and Hoi, Steven},
  booktitle={International conference on machine learning},
  pages={19730--19742},
  year={2023},
  organization={PMLR}
}

@article{metaquery,
  title={Transfer between modalities with metaqueries},
  author={Pan, Xichen and Shukla, Satya Narayan and Singh, Aashu and Zhao, Zhuokai and Mishra, Shlok Kumar and Wang, Jialiang and Xu, Zhiyang and Chen, Jiuhai and Li, Kunpeng and Juefei-Xu, Felix and others},
  journal={arXiv preprint arXiv:2504.06256},
  year={2025}
}

@misc{locdiff,
      title={LocDiff: Identifying Locations on Earth by Diffusing in the Hilbert Space}, 
      author={Zhangyu Wang and Zeping Liu and Jielu Zhang and Zhongliang Zhou and Qian Cao and Nemin Wu and Lan Mu and Yang Song and Yiqun Xie and Ni Lao and Gengchen Mai},
      year={2025},
      eprint={2503.18142},
      archivePrefix={arXiv},
      primaryClass={cs.CV},
      url={https://arxiv.org/abs/2503.18142}, 
}

@misc{lisa,
      title={LISA: Reasoning Segmentation via Large Language Model}, 
      author={Xin Lai and Zhuotao Tian and Yukang Chen and Yanwei Li and Yuhui Yuan and Shu Liu and Jiaya Jia},
      year={2024},
      eprint={2308.00692},
      archivePrefix={arXiv},
      primaryClass={cs.CV},
      url={https://arxiv.org/abs/2308.00692}, 
}

@misc{anchorseg,
      title={AnchorSeg: Language Grounded Query Banks for Reasoning Segmentation}, 
      author={Rui Qian and Chuanhang Deng and Qiang Huang and Jian Xiong and Mingxuan Li and Yingbo Zhou and Wei Zhai and Jintao Chen and Dejing Dou},
      year={2026},
      eprint={2604.18562},
      archivePrefix={arXiv},
      primaryClass={cs.CV},
      url={https://arxiv.org/abs/2604.18562}, 
}

@inproceedings{GeoToken,
  title={GeoToken: Hierarchical Geolocalization of Images via Next Token Prediction},
  author={Ghasemi, Narges and Ziashahabi, Amir and Avestimehr, Salman and Shahabi, Cyrus},
  booktitle={IEEE International Conference on Data Mining},
  year={2025}
}

@misc{zhu2021vigorcrossviewimagegeolocalization,
      title={VIGOR: Cross-View Image Geo-localization beyond One-to-one Retrieval}, 
      author={Sijie Zhu and Taojiannan Yang and Chen Chen},
      year={2021},
      eprint={2011.12172},
      archivePrefix={arXiv},
      primaryClass={cs.CV},
      url={https://arxiv.org/abs/2011.12172}, 
}

@misc{kamath2021mdetrmodulateddetection,
      title={MDETR -- Modulated Detection for End-to-End Multi-Modal Understanding}, 
      author={Aishwarya Kamath and Mannat Singh and Yann LeCun and Gabriel Synnaeve and Ishan Misra and Nicolas Carion},
      year={2021},
      eprint={2104.12763},
      archivePrefix={arXiv},
      primaryClass={cs.CV},
      url={https://arxiv.org/abs/2104.12763}, 
}

@misc{li2022groundedlanguageimagepretraining,
      title={Grounded Language-Image Pre-training}, 
      author={Liunian Harold Li and Pengchuan Zhang and Haotian Zhang and Jianwei Yang and Chunyuan Li and Yiwu Zhong and Lijuan Wang and Lu Yuan and Lei Zhang and Jenq-Neng Hwang and Kai-Wei Chang and Jianfeng Gao},
      year={2022},
      eprint={2112.03857},
      archivePrefix={arXiv},
      primaryClass={cs.CV},
      url={https://arxiv.org/abs/2112.03857}, 
}

@misc{liu2024groundingdinomarryingdino,
      title={Grounding DINO: Marrying DINO with Grounded Pre-Training for Open-Set Object Detection}, 
      author={Shilong Liu and Zhaoyang Zeng and Tianhe Ren and Feng Li and Hao Zhang and Jie Yang and Qing Jiang and Chunyuan Li and Jianwei Yang and Hang Su and Jun Zhu and Lei Zhang},
      year={2024},
      eprint={2303.05499},
      archivePrefix={arXiv},
      primaryClass={cs.CV},
      url={https://arxiv.org/abs/2303.05499}, 
}

@misc{chen2023shikraunleashingmultimodalllms,
      title={Shikra: Unleashing Multimodal LLM's Referential Dialogue Magic}, 
      author={Keqin Chen and Zhao Zhang and Weili Zeng and Richong Zhang and Feng Zhu and Rui Zhao},
      year={2023},
      eprint={2306.15195},
      archivePrefix={arXiv},
      primaryClass={cs.CV},
      url={https://arxiv.org/abs/2306.15195}, 
}

@misc{zou2022generalizeddecodingpixelimage,
      title={Generalized Decoding for Pixel, Image, and Language}, 
      author={Xueyan Zou and Zi-Yi Dou and Jianwei Yang and Zhe Gan and Linjie Li and Chunyuan Li and Xiyang Dai and Harkirat Behl and Jianfeng Wang and Lu Yuan and Nanyun Peng and Lijuan Wang and Yong Jae Lee and Jianfeng Gao},
      year={2022},
      eprint={2212.11270},
      archivePrefix={arXiv},
      primaryClass={cs.CV},
      url={https://arxiv.org/abs/2212.11270}, 
}

@misc{ren2024pixellmpixelreasoninglarge,
      title={PixelLM: Pixel Reasoning with Large Multimodal Model}, 
      author={Zhongwei Ren and Zhicheng Huang and Yunchao Wei and Yao Zhao and Dongmei Fu and Jiashi Feng and Xiaojie Jin},
      year={2024},
      eprint={2312.02228},
      archivePrefix={arXiv},
      primaryClass={cs.CV},
      url={https://arxiv.org/abs/2312.02228}, 
}

@misc{rasheed2024glammpixelgroundinglarge,
      title={GLaMM: Pixel Grounding Large Multimodal Model}, 
      author={Hanoona Rasheed and Muhammad Maaz and Sahal Shaji Mullappilly and Abdelrahman Shaker and Salman Khan and Hisham Cholakkal and Rao M. Anwer and Erix Xing and Ming-Hsuan Yang and Fahad S. Khan},
      year={2024},
      eprint={2311.03356},
      archivePrefix={arXiv},
      primaryClass={cs.CV},
      url={https://arxiv.org/abs/2311.03356}, 
}
}
\newpage

\appendix

\twocolumn[
\begin{center}
    {\LARGE \bfseries Supplementary Material\par}
    \vspace{1.5em}
\end{center}
]

\section{Training Details}
\label{app:training_details}

Table~\ref{tab:training_hparams} summarizes the key GeoBridge training hyperparameters. The entries are taken from the run configuration, with local filesystem paths omitted for anonymity.

\begin{table}[t]
\centering
\caption{Key training hyperparameters for GeoBridge.}
\label{tab:training_hparams}
\small
\setlength{\tabcolsep}{4pt}
\renewcommand{\arraystretch}{1.08}
\begin{tabular}{@{}p{0.34\linewidth}p{0.59\linewidth}@{}}
\toprule
Item & Value \\
\midrule
Base MLLM & GLOBE-Qwen2.5-VL-7B \\
Coordinate head & Frozen PLONK Riemannian-flow head \\
Training data & MP16-Pro-subset-1M \\
Evaluation data & IM2GPS3K \\
Epochs & 1 \\
Batching & 8 GPUs, batch size 4/GPU, accumulation 4 \\
Learning rate & $1\times10^{-4}$, cosine decay, warmup 1000 steps \\
1-to-$N$ expansion & $N=8$ flow samples per image condition \\
Role tokens & $M=5$: country, region, city, latitude, longitude \\
Connector & 1-layer bidirectional connector, hidden/output dim 1024 \\
Role grouping & 3 contextual roles + 2 spatial roles \\
Losses & RFM loss; country CE weight 0.05; region CE weight 0.03; city CE weight 0.02 \\
Frozen modules & MLLM backbone and PLONK/RFM head \\
Trainable modules & Role-token embeddings, connector/projection, country/city heads \\
Optimization memory & DeepSpeed ZeRO-2 + gradient checkpointing \\
\bottomrule
\end{tabular}
\end{table}
\section{1-to-\texorpdfstring{$N$}{N} Condition Expansion}
\label{app:one_to_n}

The RFM objective contains stochasticity from the sampled base point $\epsilon$ and flow time $t$. Computing the MLLM condition $c_i$ is expensive, while evaluating the frozen RFM head on additional flow samples is comparatively cheap. We therefore use 1-to-$N$ condition expansion: for each image, compute the condition once and reuse it across $N$ independent RFM training samples.

For a mini-batch $\{(x_i,y_i)\}_{i=1}^{B}$, GeoBridge first computes
\begin{equation}
    c_i = h_{\theta}(f_{\mathrm{MLLM}}(x_i)),
    \qquad i=1,\ldots,B.
\end{equation}
Then, for each $i$, sample $N$ independent pairs $(\epsilon_{i,n},t_{i,n})$. The expanded estimator is
\begin{equation}
    \widehat{\mathcal{L}}_{\mathrm{RFM}}^{(N)}
    =
    \frac{1}{BN}
    \sum_{i=1}^{B}
    \sum_{n=1}^{N}
    \left\|
    v_{\phi}(z_{i,n},t_{i,n},c_i)
    -
    u_{i,n}
    \right\|_{T_{z_{i,n}}\mathbb{S}^{2}}^{2}.
\end{equation}
The semantic auxiliary loss is computed once per image:
\begin{equation}
    \mathcal{L}
    =
    \widehat{\mathcal{L}}_{\mathrm{RFM}}^{(N)}
    +
    \sum_{r \in \mathcal{R}_{\mathrm{adm}}}\lambda_r \mathcal{L}_r^{\mathrm{CE}}.
\end{equation}
This estimator is unbiased for the RFM expectation. If the $N$ flow samples are conditionally independent given $(x_i,y_i,c_i)$, the Monte-Carlo variance of the flow-sampling component decreases approximately as $1/N$, while the MLLM forward cost remains unchanged. The trade-off is that RFM-head compute and activation memory scale linearly with $N$.

\begin{algorithm}[t]
\caption{GeoBridge training with 1-to-$N$ expansion}
\label{alg:one_to_n}
\begin{algorithmic}[1]
\REQUIRE Image batch $x_{1:B}$, targets $y_{1:B}$, semantic labels, expansion factor $N$
\STATE Run the frozen MLLM with trainable role-token embeddings and obtain role states $H_{1:B}$
\STATE Compute conditions $c_{1:B}$ and country/city logits with the connector and auxiliary heads
\STATE Compute $\mathcal{L}_{\mathrm{sem}}$ once over the $B$ images
\STATE Repeat each condition and target $N$ times: $c_{i,n}\leftarrow c_i$, $y_{i,n}\leftarrow y_i$
\STATE Sample $\epsilon_{i,n}\sim\mathrm{Unif}(\mathbb{S}^{2})$ and $t_{i,n}\sim\mathcal{U}[0,1]$
\STATE Construct $z_{i,n}$ and target tangent velocity $u_{i,n}$ using the RFM interpolant
\STATE Predict $\hat{u}_{i,n}=v_{\phi}(z_{i,n},t_{i,n},c_{i,n})$ with frozen head parameters
\STATE Compute $\widehat{\mathcal{L}}_{\mathrm{RFM}}^{(N)}
= (BN)^{-1}\sum_{i,n}\|\hat{u}_{i,n}-u_{i,n}\|^{2}_{T_{z_{i,n}}\mathbb{S}^{2}}$
\STATE Backpropagate $\widehat{\mathcal{L}}_{\mathrm{RFM}}^{(N)}+\mathcal{L}_{\mathrm{sem}}$ into role tokens and connector modules only
\end{algorithmic}
\end{algorithm}

\section{Sweep over the Expansion Factor \texorpdfstring{$N$}{N}}
\label{app:n_sweep}

\begin{figure}[t]
\centering
\includegraphics[width=0.9\linewidth]{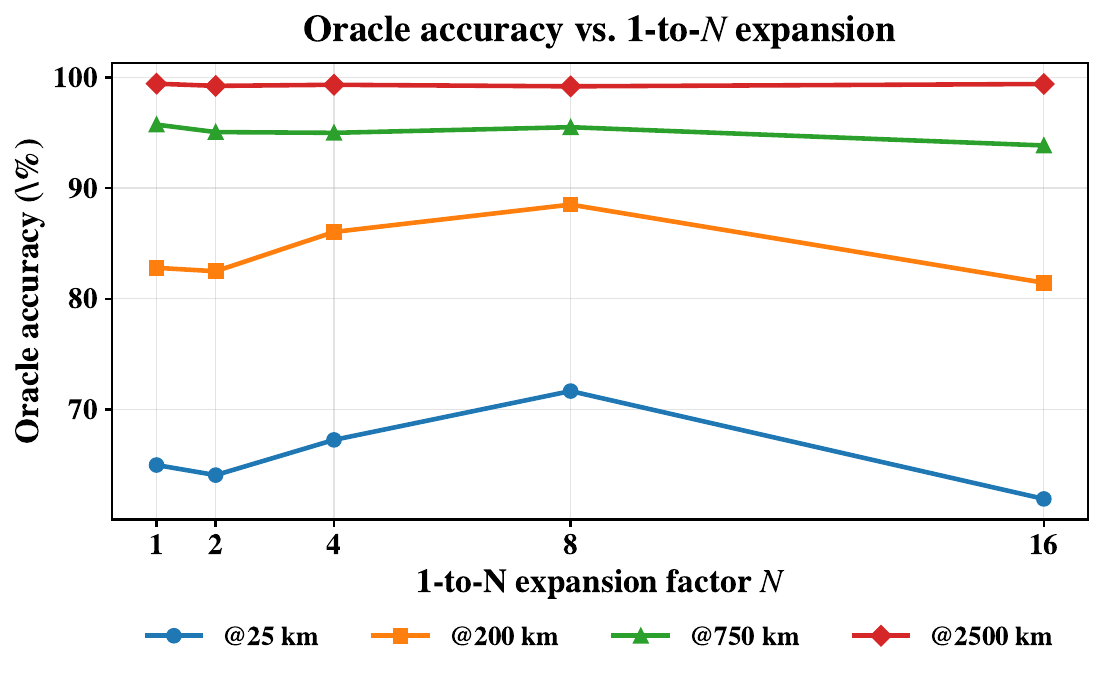}
\caption{Oracle accuracy versus 1-to-$N$ expansion.}
\label{fig:n_sweep_accuracy}
\end{figure}

Figure~\ref{fig:n_sweep_accuracy} reports an oracle-condition sweep over the expansion factor $N$ on IM2GPS3K. The oracle protocol uses teacher-forced ground-truth semantic fields and therefore probes condition-to-coordinate training rather than the deployable generated-condition setting. 

Among the strictly comparable runs, moderate expansion gives the strongest fine- and mid-range oracle accuracy: $N=8$ is best at @25 and @200. Larger expansion does not monotonically improve accuracy, suggesting that too many repeated flow samples from the same semantic condition can over-emphasize local flow-sampling variance relative to condition diversity. 

\section{Freezing versus Unfreezing the RFM Head}
\label{app:freeze_rfm}

GeoBridge freezes the pretrained RFM head and trains only the semantic interface. To verify that this is not merely a computational shortcut, we compare freezing and unfreezing the coordinate head under two settings.

\begin{figure}[t]
\centering
\includegraphics[width=\linewidth]{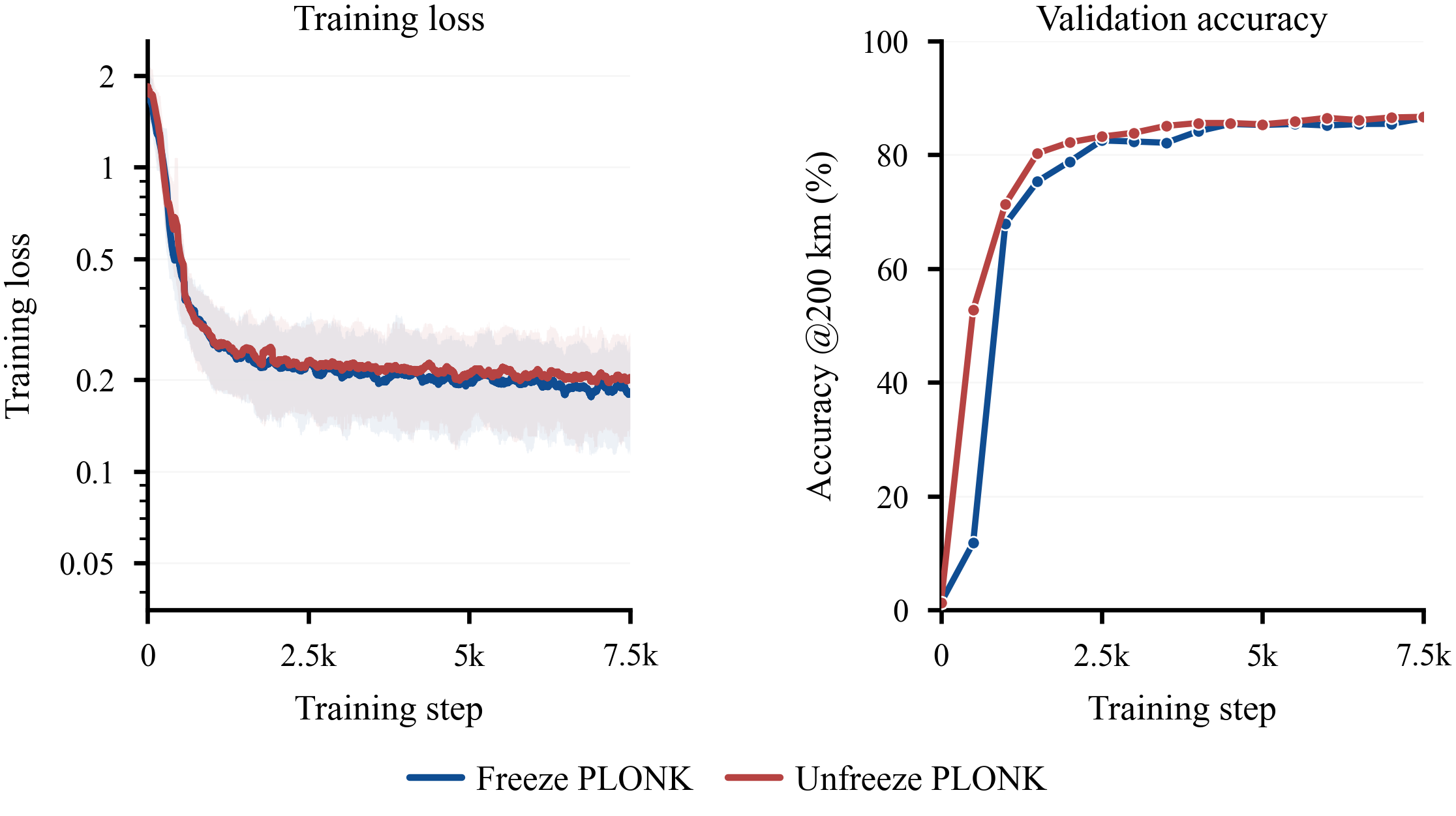}
\caption{End-to-end comparison between freezing and unfreezing the RFM head. }
\label{fig:rfm_freeze_unfreeze_e2e}
\end{figure}

\paragraph{End-to-end unfreezing from the start.}
We first train GeoBridge end-to-end with the same connector and role-token design, but allow the RFM head to update together with the bridge. As shown in Fig.~\ref{fig:rfm_freeze_unfreeze_e2e}, unfreezing the head produces nearly the same optimization curve as the frozen-head run. The training losses rapidly enter the same range, and the @200 km validation accuracy converges to an almost identical plateau. Although the unfrozen head has more trainable capacity, it does not yield a measurable accuracy gain; in mean-error evaluation, the frozen-head run is slightly better. We therefore keep the RFM head frozen in the final model.

\begin{figure}[t]
\centering
\includegraphics[width=\linewidth]{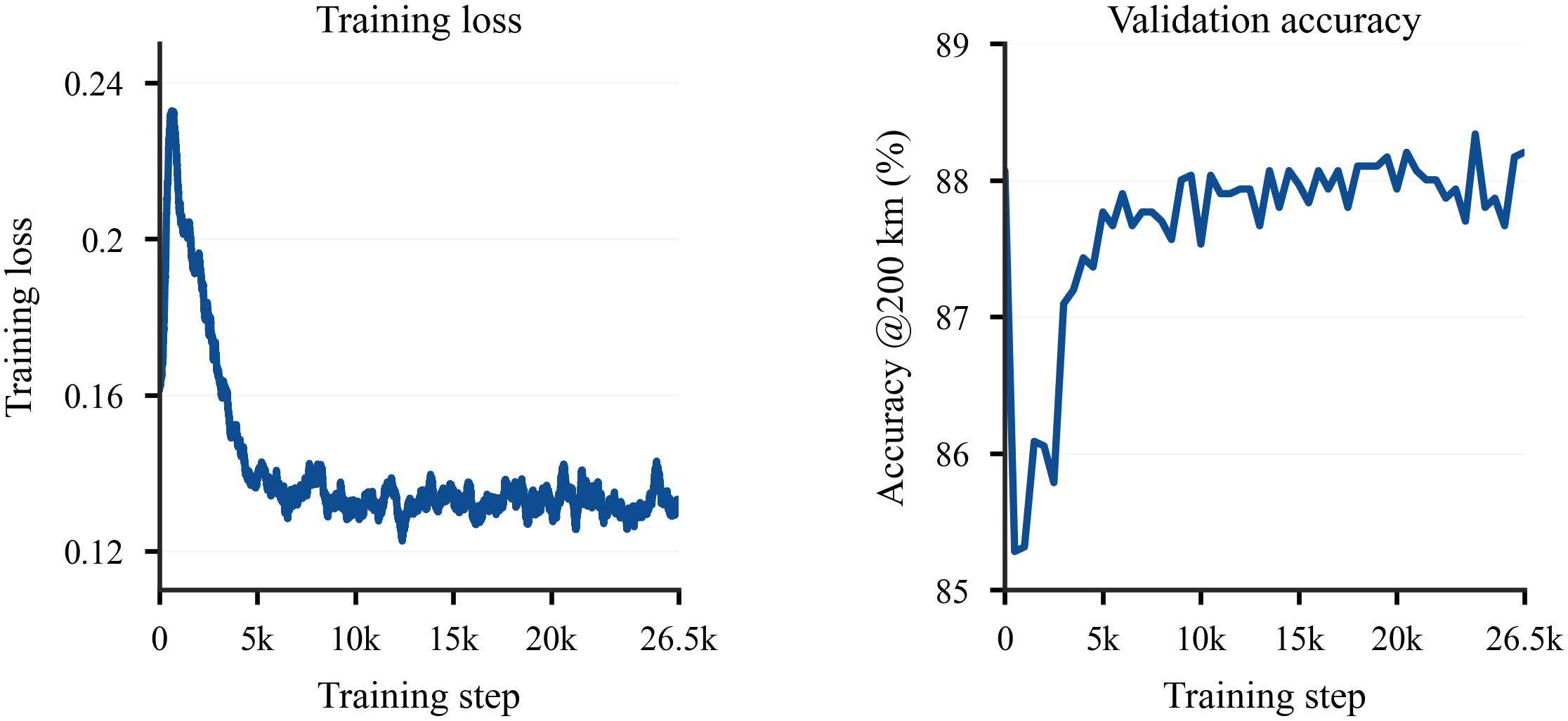}
\caption{Two-stage unfreezing after connector pretraining. We initialize from a first-stage connector trained with a frozen RFM head, then unfreeze the coordinate head for continued training. }
\label{fig:rfm_unfreeze_stage2}
\end{figure}

\paragraph{Second-stage unfreezing after connector pretraining.}
We also test a two-stage alternative: first train the GeoBridge connector with the RFM head frozen, then load the first-stage connector and continue training while unfreezing the coordinate head. Fig.~\ref{fig:rfm_unfreeze_stage2} shows that the second-stage loss first rises rapidly after unfreezing the coordinate head and then gradually stabilizes. Meanwhile, validation accuracy drops immediately after unfreezing and later recovers, but it never surpasses the first-stage frozen-head performance; longer second-stage training yields only marginal gains. This suggests that adapting the pretrained vector field after the connector has already learned its interface mainly perturbs a useful spherical decoder prior rather than improving the semantic-to-coordinate mapping.

\section{Deployable component ablation}
\label{app:deployable_ablation}
 
Table~\ref{tab:ablation} isolates each conditioning component under the \emph{oracle} condition, which measures the ceiling a variant could reach with a correct semantic prefix. Table~\ref{tab:ablation_deployable} repeats the same progression under the deployable condition---each variant using its own generated prefix on IM2GPS3K---so the two tables together separate \emph{representational ceiling} from \emph{realized gain}.
 
\begin{table}[t]
\centering
\caption{\textbf{Conditioning-interface ablation under the deployable condition (IM2GPS3K).} The same variants as Table~\ref{tab:ablation}, but each model uses its own generated semantic prefix at inference rather than a teacher-forced oracle condition. Acc@$k$ (\%); best per column in \textbf{bold}. The last row is \method's deployable result.}
\label{tab:ablation_deployable}
\small
\setlength{\tabcolsep}{3.2pt}
\begin{tabular}{lcccc}
\toprule
Conditioning variant & @25 & @200 & @750 & @2500 \\
\midrule
Single-token, MLP              & 21.72 & 42.63 & 65.34 & 82.59 \\
Role tokens, ungrouped, MLP    & 20.94 & 41.79 & 65.57 & 83.00 \\
Role tokens, grouped, MLP      & 23.36 & 48.74 & 67.89 & 84.38 \\
Role tokens, grouped, Q2Enc    & 38.50 & 52.81 & \textbf{70.90} & 84.38 \\
\textbf{Full \method{} (+ role CE)} & \textbf{38.67} & \textbf{52.89} & 70.37 & \textbf{84.42} \\
\bottomrule
\end{tabular}
\end{table}
 
Three observations follow. First, the ordering of the early steps matches the oracle study: adding ungrouped tokens does not help (it slightly dips), while grouping gives a modest gain, reaffirming that structured decomposition rather than token count is the active ingredient. Second, the connector is the dominant deployable driver: replacing the MLP with Q2Enc lifts @25 from $23.3$ to $38.5$---the single largest deployable jump, accounting for roughly $90\%$ of \method's improvement over the single-token baseline. Third, and unlike the oracle study, the auxiliary role cross-entropy is approximately neutral in deployment: it moves @25 by only $+0.2$ ($38.50{\to}38.67$) and is marginally lower at 750\,km, even though it raised the oracle ceiling by more than ten points (Table~\ref{tab:ablation}).
 
This gap is consistent with our central finding rather than at odds with it. Deployable accuracy is gated by the quality of the generated semantic prefix, so a component that sharpens how a \emph{correct} condition is exploited yields little when the condition itself is frequently wrong; role CE accordingly shows its value at the ceiling (oracle) far more than in realized accuracy (deployable). We retain it because it raises the achievable ceiling and is expected to pay off as upstream semantics improve, but we report its current deployable contribution transparently. The same reading explains why most of \method's deployable accuracy comes from the connector, which improves the semantics-to-coordinate mapping regardless of whether the underlying prefix is correct.

\section{Flow Matching and Riemannian Flow Matching}
\label{app:fm_rfm}

This section expands the coordinate generator used by GeoBridge. The head itself is inherited from RFM~\cite{chen2024flow} and kept frozen; GeoBridge only learns the semantic condition supplied to this head. We use the prior-to-data convention below, matching the sampling direction in GeoBridge: a random base point is drawn at $t=0$ and transported to a predicted geographic point at $t=1$. This is equivalent to the data-to-noise convention often used in flow-matching papers after the time change $t \mapsto 1-t$.

\paragraph{Coordinates on the sphere.}
A latitude--longitude pair $(\varphi,\lambda)$ is represented as a unit vector
\begin{equation}
    y(\varphi,\lambda)
    =
    \big(\cos\varphi\cos\lambda,\;
         \cos\varphi\sin\lambda,\;
         \sin\varphi\big)
    \in \mathbb{S}^{2}\subset\mathbb{R}^{3}.
\end{equation}
The geodesic distance between two unit vectors $p,q\in\mathbb{S}^{2}$ is
\begin{equation}
    d_{\mathbb{S}^{2}}(p,q)
    =
    \arccos\big(\mathrm{clip}(p^{\top}q,-1,1)\big),
\end{equation}
and the distance in kilometers is obtained by multiplying by the Earth radius $R_{\oplus}$.

\paragraph{Exponential and logarithmic maps.}
For $p,q\in\mathbb{S}^{2}$, let $\alpha=d_{\mathbb{S}^{2}}(p,q)$. The logarithmic map sends $q$ to a tangent vector at $p$:
\begin{equation}
    \log_{p}(q)
    =
    \frac{\alpha}{\sin\alpha}\big(q-\cos\alpha\,p\big)
    \in T_{p}\mathbb{S}^{2}.
\end{equation}
For a tangent vector $v\in T_{p}\mathbb{S}^{2}$, the exponential map is
\begin{equation}
    \exp_{p}(v)
    =
    \cos\|v\|\,p
    +
    \sin\|v\|\frac{v}{\|v\|}.
\end{equation}
In implementation, the usual small-angle limits are used when $\alpha$ or $\|v\|$ is close to zero. A vector $a\in\mathbb{R}^{3}$ can be projected to the tangent plane at $p$ by
\begin{equation}
    \Pi_{p}(a)=a-(a^{\top}p)p.
\end{equation}

\paragraph{Euclidean flow matching.}
Let \textbf{$c$} denote the GeoBridge condition, $y\in\mathbb{S}^{2}$ the target coordinate, and $\epsilon$ a base sample. In Euclidean flow matching, the interpolant is defined in $\mathbb{R}^{3}$:
\begin{equation}
    z_{t}^{\mathrm{E}}
    =
    (1-\kappa(t))\epsilon + \kappa(t)y,
    \qquad
    \kappa(0)=0,\quad \kappa(1)=1.
\end{equation}
The target velocity is
\begin{equation}
    u_{t}^{\mathrm{E}}
    =
    \frac{d z_{t}^{\mathrm{E}}}{dt}
    =
    \dot{\kappa}(t)(y-\epsilon),
\end{equation}
and the Euclidean FM loss is
\begin{equation}
    \mathcal{L}_{\mathrm{FM}}
    =
    \mathbb{E}_{(y,c),\,\epsilon,\,t}
    \left[
        \left\|
        v_{\phi}(z_{t}^{\mathrm{E}},t,c)
        -
        u_{t}^{\mathrm{E}}
        \right\|_{2}^{2}
    \right].
\end{equation}
Because $z_{t}^{\mathrm{E}}$ does not generally stay on $\mathbb{S}^{2}$, Euclidean FM requires a final projection or normalization to produce a valid location.

\paragraph{Riemannian flow matching on $\mathbb{S}^{2}$.}
RFM instead constructs the interpolation directly on the sphere. Given $\epsilon\sim\mathrm{Unif}(\mathbb{S}^{2})$ and target $y\in\mathbb{S}^{2}$, define
\begin{equation}
    z_{t}
    =
    \exp_{\epsilon}
    \left(
        \kappa(t)\log_{\epsilon}(y)
    \right)
    \in \mathbb{S}^{2}.
\end{equation}
The corresponding tangent velocity is
\begin{equation}
    u_{t}
    =
    \dot{\kappa}(t)\,
    \mathrm{PT}_{\epsilon\rightarrow z_{t}}
    \left[
        \log_{\epsilon}(y)
    \right]
    \in T_{z_t}\mathbb{S}^{2},
\end{equation}
where $\mathrm{PT}_{\epsilon\rightarrow z_{t}}$ denotes parallel transport along the geodesic from $\epsilon$ to $z_t$. The RFM objective is
\begin{equation}
    \mathcal{L}_{\mathrm{RFM}}
    =
    \mathbb{E}_{(y,c),\,\epsilon,\,t}
    \left[
        \left\|
        v_{\phi}(z_t,t,c)
        -
        u_t
        \right\|_{T_{z_t}\mathbb{S}^{2}}^{2}
    \right].
\end{equation}
At inference, the frozen RFM head transports a random point on the sphere by integrating
\begin{equation}
    \dot{z}_{t}=v_{\phi}(z_t,t,c),
    \qquad
    z_0\sim\mathrm{Unif}(\mathbb{S}^{2}),
\end{equation}
with manifold Euler steps
\begin{equation}
    z_{t+\Delta t}
    =
    \exp_{z_t}
    \left(
        \Delta t\,v_{\phi}(z_t,t,c)
    \right),
    \qquad
    \Delta t=1/K.
\end{equation}
This keeps every intermediate point on $\mathbb{S}^{2}$ and avoids the Euclidean projection mismatch.

\section{Additional Qualitative Results}
\label{app:qualitative_buckets}

\begin{figure*}[t]
\centering
\includegraphics[width=\textwidth]{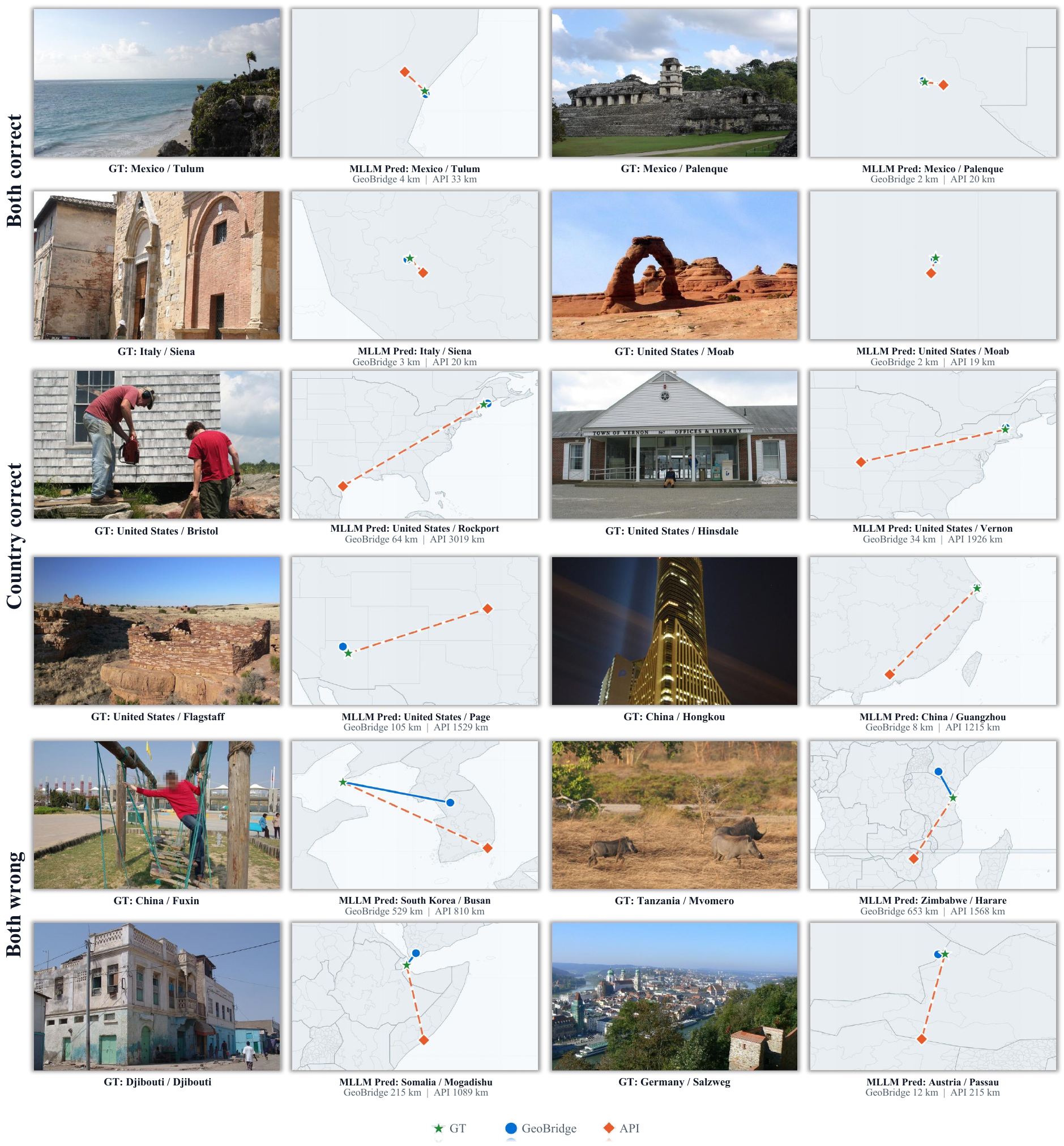}
\caption{Additional qualitative results in the \emph{both correct, country correct, city wrong, both wrong} bucket.}
\label{fig:qual_both_correct_more}
\end{figure*}

The main paper groups deployable predictions by whether the generated semantic condition contains the correct country and city. Figure \ref{fig:qual_both_correct_more} provides additional examples for each bucket to show how the coordinate decoder responds to semantic condition quality.

\section{Inference efficiency}
\method\ is attached after the same frozen GLOBE semantic-generation pass used by the place-name baseline, so it does not add another autoregressive decode. Consequently, the MLLM-side TTFT and TPOT are unchanged: \method\ only incurs a fixed post-generation cost for reading the role-token states, projecting them into a single condition, and running the frozen RFM sampler. In the current implementation, the role-token readout is profiled as an uncached one-forward pass and takes $0.62$s median latency, while the $32$-step RFM sampler adds $0.25$s; together they form a conservative $0.85$s median bridge overhead. The deterministic linear-layer MAC estimate is $2.33$G, dominated by the RFM head. A deployable implementation can reuse the generated-prefix KV cache for the role-token readout, so the measured latency should be interpreted as an implementation upper bound rather than a required extra reasoning pass.

\section{Discussion and Limitations}
 
\paragraph{What limits deployable accuracy.}
Our analysis consistently localizes the limit to the condition rather than the head. Under an oracle condition the frozen RFM head reaches a far higher ceiling; deployable error tracks the correctness of the generated semantics; and stronger geographic backbones yield better conditions. Together these reframe the open problem from designing a better coordinate head to producing a more distribution-aligned geospatial condition. That goal is complementary to advances in MLLM reasoning: a reasoning- or CoT-augmented backbone would improve the very condition \method\ decodes, so the two lines compose rather than compete.
 
 
\paragraph{Formatted prompting.}
\method\ reads its condition from a fixed prompt template, with the role tokens occupying designated positions. It therefore does not yet support coordinate estimation embedded in free-form generation---for example, emitting a location partway through open-ended reasoning. A natural extension is to fold the learnable role tokens into the MLLM's own vocabulary, so the model can emit them directly and produce flexible, on-demand coordinate estimates under arbitrary outputs.
 
\paragraph{Sensitivity to the textual condition.}
Because we inject ground-truth semantic text during training, the learned condition is sensitive to surface form: different verbalizations of the same place---abbreviations, alternate names, or language variants---can yield markedly different predictions, since coordinates are bound to specific training-set strings rather than to a normalized geographic entity. Normalizing place references, or grounding them in an entity vocabulary, is a promising way to reduce this variance.
 

\section{Training Prompt and Teacher-Forced Prefix}
\label{app:training_prompt}

GeoBridge uses structured semantic prompts to expose geographic fields before reading the learned role-token states. We used two prompt templates during development.

\paragraph{Country--state--city prompt.}
We also used a more structured administrative prompt that asks for three geographic fields:
\begin{quote}
\small\ttfamily\noindent
You must analyze the input image and provide a structured location prediction at exactly four levels of geographic granularity:\\
\hspace*{1em}1. Country\\
\hspace*{1em}2. State (Administrative region)\\
\hspace*{1em}3. City (e.g., "Auschwitz", "Golden Gate Bridge", "Forbidden City")
\end{quote}
The corresponding teacher-forced answer prefix is
\begin{quote}
\small\ttfamily\noindent
\{\{"country": "\{country\_text\}", "state": "\{region\_text\}", "city": "\{city\_text\}"<coordinates>
\end{quote}

\paragraph{Country--city prompt.}
The first template follows the GLOBE-style semantic geolocation format and asks the MLLM to output only country and city fields:
\begin{quote}
\small\ttfamily\noindent
Based on the provided image, please output its location (country and city) directly without any explanation.\\
Your final answer includes these two lines in your response:\\
country: [country name]\\
city: [city name]
\end{quote}
For image $i$, the teacher-forced answer prefix is constructed as
\begin{quote}
\small\ttfamily\noindent
teacher\_answer = f"country: \{country\_text\} city: \{city\_text\}<coordinates>"
\end{quote}

In both templates, \verb|<coordinates>| is a data-formatting placeholder rather than a literal text span used by the language model. Before forwarding through the MLLM, this placeholder is replaced by $M$ learnable special tokens. 

For semantic supervision, we apply a field-level mask to handle incomplete administrative annotations. 
If the annotation file does not contain a valid region/state entry for a sample, the corresponding region CE term is set to zero for that sample. 
More generally, when a region head is enabled, the semantic loss can be written as
\begin{equation}
    \mathcal{L}_{\mathrm{sem}}
    =
    \lambda_{\mathrm{cty}} \mathcal{L}^{\mathrm{CE}}_{\mathrm{cty}}
    +
    \lambda_{\mathrm{reg}} m_{\mathrm{reg}} \mathcal{L}^{\mathrm{CE}}_{\mathrm{reg}}
    +
    \lambda_{\mathrm{city}} \mathcal{L}^{\mathrm{CE}}_{\mathrm{city}},
\end{equation}
where $m_{\mathrm{reg}}\in\{0,1\}$ indicates whether a valid region/state label is available. 
For GLOBE-based GeoBridge, the base model predicts country and city but not region; therefore we keep $\lambda_{\mathrm{reg}}=0$ throughout. 
The region token is still retained as a contextual role, but it is not optimized with a region classification loss in this setting.

\end{document}